\documentclass[preprint,12pt]{elsarticle}

\usepackage{tcolorbox}
\tcbuselibrary{skins}
\usepackage{xcolor}
\usepackage[utf8]{inputenc}
\usepackage{graphicx, float, subcaption, caption}
\usepackage{amsmath, amssymb, bm, mathtools}
\usepackage{leftidx, color, xcolor}
\usepackage{hyperref}
\usepackage{algorithm, algpseudocode}
\usepackage{ragged2e}
\usepackage{times}
\usepackage{natbib}
\usepackage{titlesec}
\usepackage{lineno}
\usepackage{amsmath}
\usepackage{newunicodechar}
\newunicodechar{∼}{\textasciitilde}
\linenumbers

\usepackage{algorithm}
\usepackage{algpseudocode}
\usepackage{amsmath}

\usepackage{url}          
\usepackage{hyperref}
\hypersetup{
  colorlinks=true,
  urlcolor=blue,
  citecolor=blue,
  linkcolor=blue
}
\usepackage{mathrsfs}  

\DeclareSymbolFont{newfont}{OML}{cmm}{m}{it}
\DeclareMathSymbol{\Varrho}{3}{newfont}{37}

\usepackage{setspace}
\usepackage{titlesec}
\titleformat{\paragraph}[runin]
  {\normalfont\normalsize\bfseries}
  {\theparagraph}{1em}{}
\titlespacing*{\paragraph}{0pt}{3.25ex plus 1ex minus .2ex}{1em}
\journal{Results in Engineering}

\usepackage{titlesec}
\begin{document}
\nolinenumbers
\begin{frontmatter}
\title{Resilient Decentralized Radar-LiDAR-Inertial Odometry in Challenging Subterranean Environments}
\author[mainaddress]{Moumita Mukherjee\corref{correspondingauthor}}
\author[mainaddress]{Avijit Banerjee}
\author[mainaddress]{Anton Koval}
\author[mainaddress]{George Nikolakopoulos}

\cortext[correspondingauthor]{Corresponding author: moumita.mukherjee@ltu.se}

\address[mainaddress]{Robotics and AI Team, Department of Computer Science, Electrical Engineering and Space Engineering, Luleå University of Technology, Luleå, Sweden}

\begin{abstract}
Accurate localization in GPS-denied and visually degraded environments affected by smoke, dust, or fog remains a critical challenge for autonomous systems. While LiDAR provides high-resolution spatial perception, its performance deteriorates under particulate occlusion. In contrast, frequency-modulated continuous-wave (FMCW) radar offers robustness under such conditions but suffers from sparse measurements and outliers. This article presents a resilient decentralized LiDAR Radar Inertial Odometry (LRIO) framework that fuses the complementary strengths of LiDAR, radar, and inertial sensing to achieve robust pose estimation in challenging subterranean environments. At the core of the framework is an Adaptive Residual Thresholding (ART) mechanism that autonomously detects temporal sensor degradation and selectively adjusts the fusion process to minimize unreliable measurements. Building on this, a Robust Error-Tolerant Optimal Information Filter (RETOIF) is employed within a decentralized fusion architecture to integrate LiDAR-inertial and multiple radar-inertial odometry estimates in a fault-aware manner, enhancing resilience to environmental disturbances and intermittent sensor failures. The proposed framework is evaluated using real-world sensor data collected in GPS-denied subterranean tunnels under smoke-free and smoke-degraded conditions. In the smoke-free experiment, the proposed RETOIF framework achieves a root mean square localization error (RMSE) of 2.1733, compared with 2.9929 for the conventional centralized LRIO approach, corresponding to a 27.4\% reduction in RMSE. Experimental results further demonstrate the framework's ability to maintain pose estimation under smoke-induced sensor degradation, highlighting its potential for resilient autonomous navigation and mapping in challenging subterranean environments.
\end{abstract}
\end{frontmatter}
\section{Introduction} 
Following the DARPA Subterranean Challenge~\texorpdfstring{\cite{lindqvist2022compra, agha2021nebula}}{[1,2]}, autonomous navigation in subterranean (Sub-T) environments has become a key research focus, driving advances in perception, localization, and mapping under extreme conditions. Sub-T settings, such as lava tubes, mines, and tunnels, lack GPS and often impair visual sensors due to low light. While LiDAR and vision based methods have enabled accurate localization in many cases~\texorpdfstring{\cite{taleb2025automotive}}{[3]}, their performance degrades severely in dense smoke, dust, fog, and precipitation~\texorpdfstring{\cite{kim2023empirical, piroli2023energy}}{[4,5]}. These environmental factors limit the reliability of LiDAR in real-world Sub-T deployments~\texorpdfstring{\cite{park2025lidar, vargas2020weather}}{[6,7]}. FMCW radars offer a promising complementary modality, providing robust and consistent performance in poor visibility. Motivated by the complementary strengths of LiDAR and radar, this work presents a decentralized LiDAR Radar Inertial Odometry (LRIO) framework designed for resilient localization in GPS-denied and visually degraded environments. The proposed framework minimizes the limitations of individual sensors, enhances robustness against environmental degradation, and enables reliable odometry and mapping for autonomous robotic operation in challenging Sub-T environments.
\subsection{Related Work}
\begin{figure*}[h]
\centering
\vspace{-0.5mm}
\includegraphics[width=1.0\textwidth]{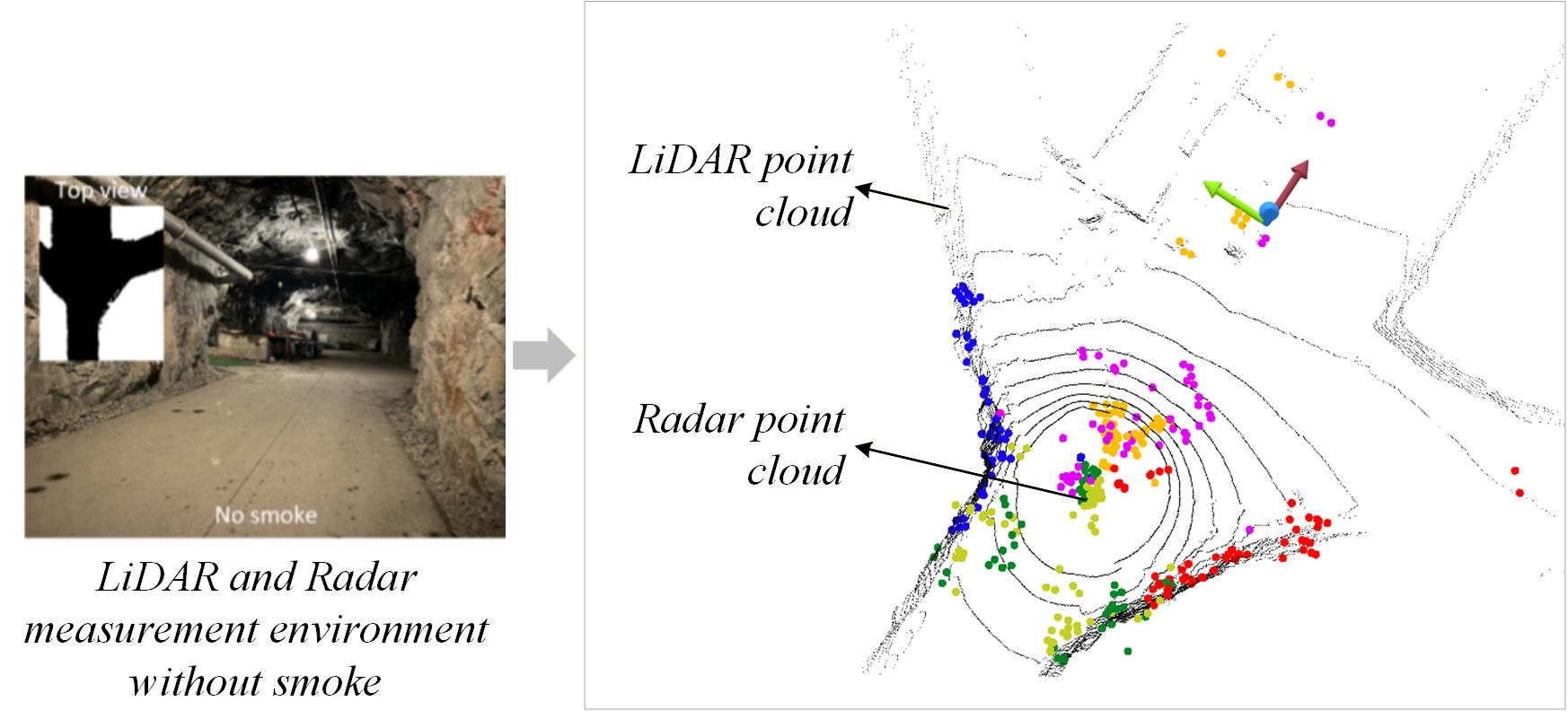}
\caption{The performance of a 3D Velodyne LiDAR and multiple millimetre-wave radars was evaluated in a GPS-denied Sub-T tunnel without smoke. LiDAR scans (black points) clearly represent the tunnel structure, while radar points (multi-coloured, each representing a different sensor) are sparse and lack discernible structure.}
\label{fig: Raw_scans_pointclouds_LiDAR_Radar}
\end{figure*}

Compared to the dense, structured, and low-noise observations obtained from LiDAR, radar sensors yield measurements that are sparse, noisy, locally clustered, and prone to flickering frequently. An example of this distinction is illustrated in Fig.~\ref{fig: Raw_scans_pointclouds_LiDAR_Radar}. These radars provide range and Doppler velocity measurements only when relative speed exists between the autonomous vehicle and its surroundings. Moreover, radar point clouds are susceptible to multipath reflections and dynamic objects such as pedestrians~\cite{chen2023drio}, which introduce outliers into the measurements. Consequently, ego-velocity estimation and relative pose derived from radar data can exhibit substantial drift~\cite{doer2020ekf, zhuang20234d, doer2021yaw} influenced by factors such as sensor placement, highly dynamic environments~\cite{chen2023drio}, and reflections from surfaces like walls and floors. Despite these challenges, recent progress in radar perception has led to notable improvements in ego velocity and pose estimation, dynamic target tracking, and environmental mapping.
\subsubsection{Radar Inertial Odometry}

In the related community, various types of radar sensors have already been utilized for localization, each producing unique characteristics of the point cloud. These include Frequency Modulated Continuous Wave (FMCW) radars~\cite{herraez2024radar},\cite{nissov2024degradation}, Doppler radars~\cite{barlow1949doppler}, advanced 4D automotive radars~\cite{zhuang20234d},~\cite{ doer2021yaw},~\cite{choi2023msc} and 4D imaging radars~\cite{zhang20234dradarslam}. Among these, 4D imaging radars~\cite{zhuang20234d},~\cite{zhang20234dradarslam} often produce rich point cloud structures, facilitating pose estimation through scan matching algorithms like ICP. The estimated relative pose, obtained from scan matching algorithms, combined with velocity estimates and IMU measurements, is integrated into an Extended Kalman Filter (EKF) to derive odometry information. The study in~\cite{doer2020ekf} suggests that a single EKF can effectively estimate pose from raw point clouds from radar sensors. Moreover, they implicitly assume smooth, reflective environments, which limits their robustness in challenging environments like Sub-T. Thus, in our previous work for RIO~\cite{mukherjee2026resilient} the field deployment and datasets obtained from continuous IWR6843AoP millimetre-wave radars faced difficulties in achieving reliable pose estimation using a single EKF. Since, these radars produce point clouds with a lack of discernible structure, depicted in Fig.~\ref{fig: Raw_scans_pointclouds_LiDAR_Radar} and consist of only a few scattered points moving around the robot. That sparsity renders scan-matching algorithms ineffective, leading to complete dead-reckoning with substantial drift in such scenarios. The proposed MRIO algorithm~\cite{mukherjee2026resilient}presents a novel two-stage Extended Kalman Filtering (EKF) framework, designed as a loosely coupled architecture.

\subsubsection{LiDAR Radar Inertial Odometry}
\begin{figure*}[h]
\centering
\includegraphics[width=1.0\textwidth]{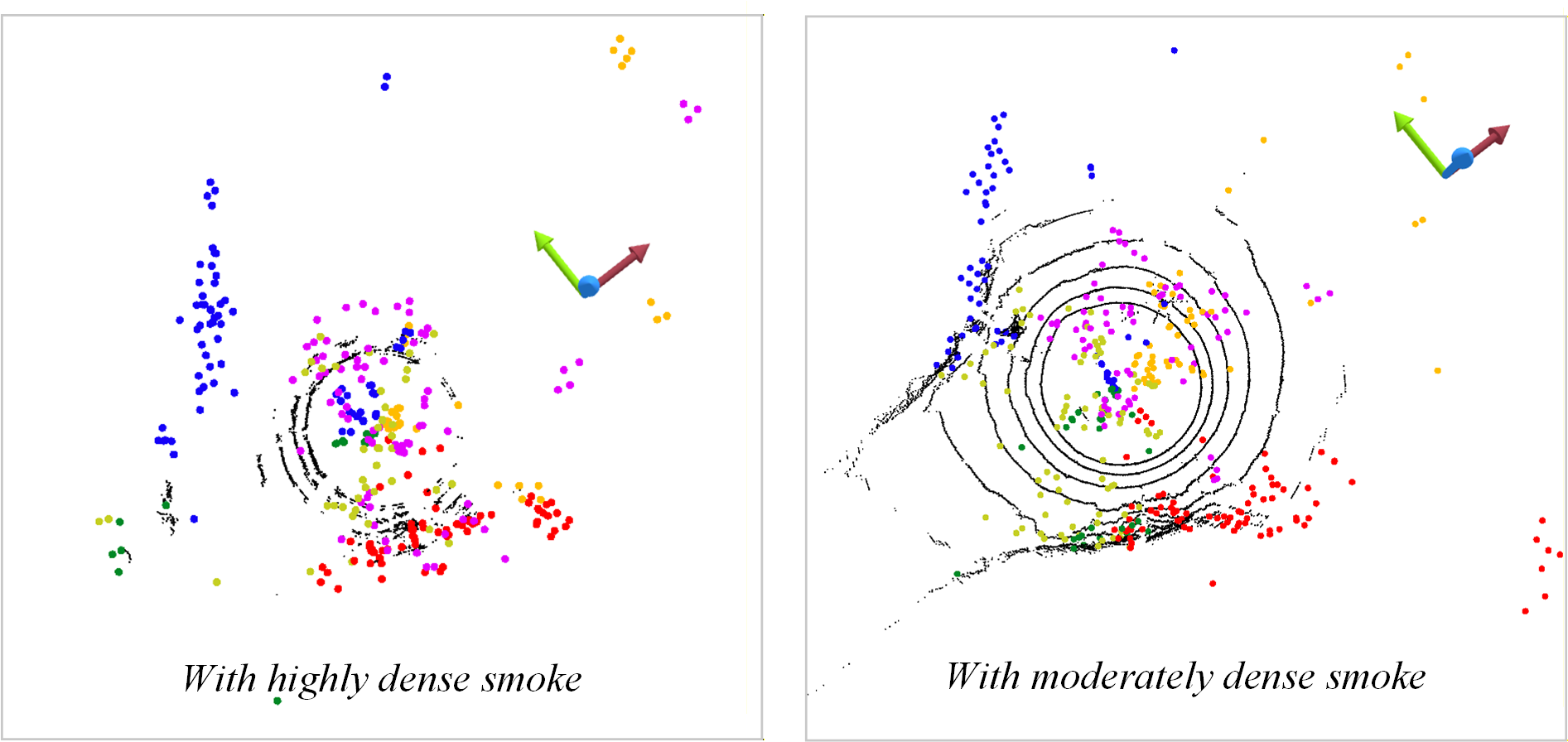}
\caption{Performance evaluation of 3D Velodyne LiDAR and IWR6843AoP mmWave radars in GPS-denied tunnels under fog shows severe LiDAR degradation, while radar remains unaffected}
\label{fig:In_smoke_scans_pointclouds_LiDAR_radar}
\end{figure*}

The performance of Radar Inertial Odometry (RIO) is strongly affected by radar mounting geometry, field of view, and environmental conditions such as pedestrian movements, dynamic scenes, glass doors and windows, and reflective floors, all of which can degrade odometry accuracy. To overcome this drift during unknown tunnel exploration, multi-sensor fusion becomes imperative. LiDAR remains a standard modality for 3D perception owing to its accurate range measurements~\cite{teng2024multi}, yet its reliability declines significantly in smoke, fog, and similar adverse conditions, as demonstrated in Fig.~\ref{fig:In_smoke_scans_pointclouds_LiDAR_radar}. In contrast, FMCW radars maintain stable performance in dust, fog, rain, and snow due to their longer wavelengths, making them a robust complement to LiDAR. Consequently, combining LiDAR and radar enables more resilient localization and mapping by exploiting the complementary strengths of both modalities. 
The LRIO framework in~\cite{nissov2024degradation} uses a sliding-window factor graph to fuse LiDAR, radar, and IMU data, improving robustness through tightly coupled multimodal fusion. However, its batch optimization can introduce latency, limiting responsiveness in dynamic or degraded environments. The continuous-time LiDAR radar inertial odometry in~\cite{li2024continuous} tightly integrates LiDAR point to plane constraints, raw IMU data, and radar Doppler measurements for improved outlier resilience, but this tightly coupled formulation increases computational cost, and its coarse to fine initialization may struggle under poor initial estimates or sensor degradation. Super Odometry~\cite{zhao2021super} provides an IMU centric pipeline with a dynamic octree for efficient point organization and strong performance in challenging conditions, yet it remains sensitive to IMU drift and incurs additional computational overhead. The centralized EKF-based LiDAR radar fusion in~\cite{kim2020extended} is simpler but prone to drift under adverse weather or LiDAR degradation due to its lack of adaptive degradation handling. To address these limitations, the proposed LRIO based on the RETOIF framework performs classical detection through residual generation, and adaptive thresholding for detection of outliers and eliminates these outliers through the FROIF framework~\cite{mukherjee2021decentralized}. This enables reliable selection of the most probable measurements for downstream fusion, resulting in accurate and robust pose estimation.
\subsubsection{Centralized vs Decentralized Multi Sensor Fusion}
Robust state estimation in multi-sensor fusion requires effective handling of sensor anomalies. While centralized fusion mitigates persistent anomalies via covariance tuning, it cannot adapt to temporal sensor failures. This is evident in LiDAR–Inertial Odometry (LIO), where dense smoke causes localized and temporary degradation in mapping and localization. Due to its limited ability to rapidly adjust covariance during such short-term disruptions, centralized fusion is less effective. In contrast, decentralized fusion dynamically adapts covariance based on sensor reliability, enabling robust handling of transient failures. Prior works~\cite{sun2003multi,yoon2023adaptive,mukherjee2021decentralized,mukherjee2022resilient,mukherjee2023fault,mukherjee2022auto,mukherjee2024autonomous,mukherjee2023decentralized} consistently demonstrate its superior resilience and performance.
\subsubsection{Autonomous Thresholding}
In decentralized fusion~\cite{mukherjee2021decentralized,sun2004multi,noack2017decentralized}, failure detection and isolation rely on thresholding to identify sensor-induced inaccuracies. Adaptive thresholding enables real-time performance monitoring, distinguishing nominal from degraded behaviour and autonomously rejecting inaccurate measurements. Multi-layer fusion architectures~\cite{sun2004multi} further enhance reliability, with early layers performing complementary detection~\cite{mukherjee2023fault,mukherjee2022auto} and subsequent layers isolating faults~\cite{mukherjee2021decentralized}. Thresholding thus remains central to robust anomaly detection and decision-making. A major limitation of fixed, preconfigured threshold based schemes is their assumption that optimal operational thresholds are known in advance. In practice, such predefined values are often suboptimal, as no single constant threshold can reliably accommodate all scenarios across a dataset. Even with well-chosen initial thresholds, continuous adaptation is required to reflect changing conditions of system dynamics. Most decentralized fusion methods rely on covariance based metrics derived from the innovation vector of the Kalman filtering~\cite{pirmoradi2009fault,yuan2021fault}.  Although these metrics capture changes caused by sensor degradation, they cannot determine the temporal duration of failure, a limitation noted in prior work~\cite{mukherjee2023fault,mukherjee2022auto,chafouk2011fault}. Another Residue analysis algorithms typically use statistical tests such as the t-test or Chi-square tests. However, these statistical methods depend on batch processing. Rather than relying on fixed thresholds derived from historical data, the proposed decentralized LiDAR radar fusion approach employs adaptive thresholding to assess sensor degradation through statistical evaluation of residuals. The Adaptive Residual Thresholding (ART) framework determines thresholds in real time using instantaneous residual information, without dependence on accumulated data. Consequently, the proposed Robust Error-Tolerant Optimal Information Filter (RETOIF) is well-suited for online implementation, eliminating reliance on batch processing and reducing latency associated with conventional statistical methods.
\subsection{Contributions}
In contrast to existing state of the art methods, this article presents a novel Resilient Complementary fusion LRIO based localization, enabling robust pose estimation in GPS-denied, visually degraded subterranean environments. By tightly fusing complementary sensing modalities. The proposed framework maintains accurate localization, particularly under severe smoke conditions where LiDAR reliability degrades. The key contributions of this work are as follows:
\begin{itemize}
    \item This article proposes a decentralized multi-sensor fusion architecture that integrates LiDAR inertial odometry from a 3D Velodyne LiDAR and four heterogeneous FMCW TI IWR6843AOP radars using the RETOIF framework. By fusing these five complementary sources of measurement with different modalities in a decentralized manner, the proposed RETOIF algorithm improves pose accuracy and robustness through cross-modal sensing synergy.
    \item A novel adaptive thresholding scheme is introduced to autonomously detect temporal sensor degradation using the residual of maximum likelihood state estimates. 
    \item In addition to providing reliable pose estimates for autonomous navigation, the proposed LRIO framework also enables the construction of a global map by transforming radar body-frame point clouds into a global reference frame.
    \item For performance analysis of the proposed method, a series of representative experiments in real Subterranean terrain were conducted using a Pioneer 3-AT ground robotic platform equipped with LiDAR, radar, and IMU sensors. The platform was tested in three distinct scenarios within a GPS-denied subterranean environment: (a) No smoke scenario (b) mild smoke scenario, and (c) dense smoke scenario. All experiments were carried out along identical trajectories within the same environment to ensure consistent conditions for assessing the effectiveness of the proposed algorithm. Experimental results and comparison with centralized fusion have demonstrated the effectiveness of the proposed decentralized approach.
\end{itemize}
The remainder of this article is organized as follows: Section $2$ introduces the notation used throughout the article. Section $3$ presents the problem formulation, while Section $4$ describes the proposed Robust Error-Tolerant Optimal Information Filter (RETOIF) algorithm and its application to LiDAR Radar Inertial Odometry (LRIO) in challenging subterranean (Sub-T) environments. Section $5$ presents the experimental evaluation and results, and Section $6$ concludes the article.
\section{Notation}
The system state and its estimate at time step $k$ are given by $\mathscr{X}_k, \hat{\mathscr{X}}_k \in \mathbb{R}^{n_s}$. The process and measurement models are defined as $f:\mathbb{R}^{n_s} \rightarrow \mathbb{R}^{n_s}$ and $h:\mathbb{R}^{n_s} \rightarrow \mathbb{R}^{m}$. The measurement set from $n$ sensor modalities is $y=[y_1,\dots,y_n]\in\mathbb{R}^{m\times n}$, where $y_i\in\mathbb{R}^{m}$. The $n$ leave one out likelihood estimators, each incorporating measurements from $(n-1)$ sensor modalities, are defined as $\hat{\mathscr{X}}_{b_i}=f_{b_i}(y\setminus y_i)$, where $y\setminus y_i$ denotes the measurement matrix excluding its $i$th column, with $f_{b_i}:\mathbb{R}^{m\times(n-1)}\rightarrow\mathbb{R}^{n_s}$, $i=1,\dots,n$. The centralized estimator, incorporating all $n$ sensor measurements, is defined as $\hat{\mathscr{X}}_n=f_n(y_1,\dots,y_n)$. The residuals are given by $\mathfrak{R}_{f_i} = \hat{\mathscr{X}}_{n} - \hat{\mathscr{X}}_{b_i}$, $i=1,\dots,n$, and the stacked residual matrix is $\mathfrak{R}_e = [\mathfrak{R}_{f_1} \;\cdots\; \mathfrak{R}_{f_n}]^T \in \mathbb{R}^{n \times 2}$. The adaptive threshold region is defined as $\Delta_k \in [\Delta_k^{-}, \Delta_k^{+}]$, where $\Delta_k^{-}, \Delta_k^{+} \in \mathbb{R}\geq 0$ and $\mathcal{K} \in \mathbb{R}^+$ is a tuning parameter. The logical selection variable is $\delta_{l_k} \in \{0,1\}$. 
The fused estimation-error covariance is denoted by
$P_k\in\mathbb{R}^{n_s\times n_s}$, while
the stacked covariance matrix is defined as
$\Sigma_k=\operatorname{blkdiag}
(\Sigma_{1,k},\dots,\Sigma_{n,k})$,
assuming mutually uncorrelated estimation errors.
The optimal stacked fusion-weight matrix is denoted
by $W_k\in\mathbb{R}^{nn_s\times n_s}$.

\section{Problem Formulation}
\begin{figure}[h]
\centering
\includegraphics [width=0.5\textwidth]{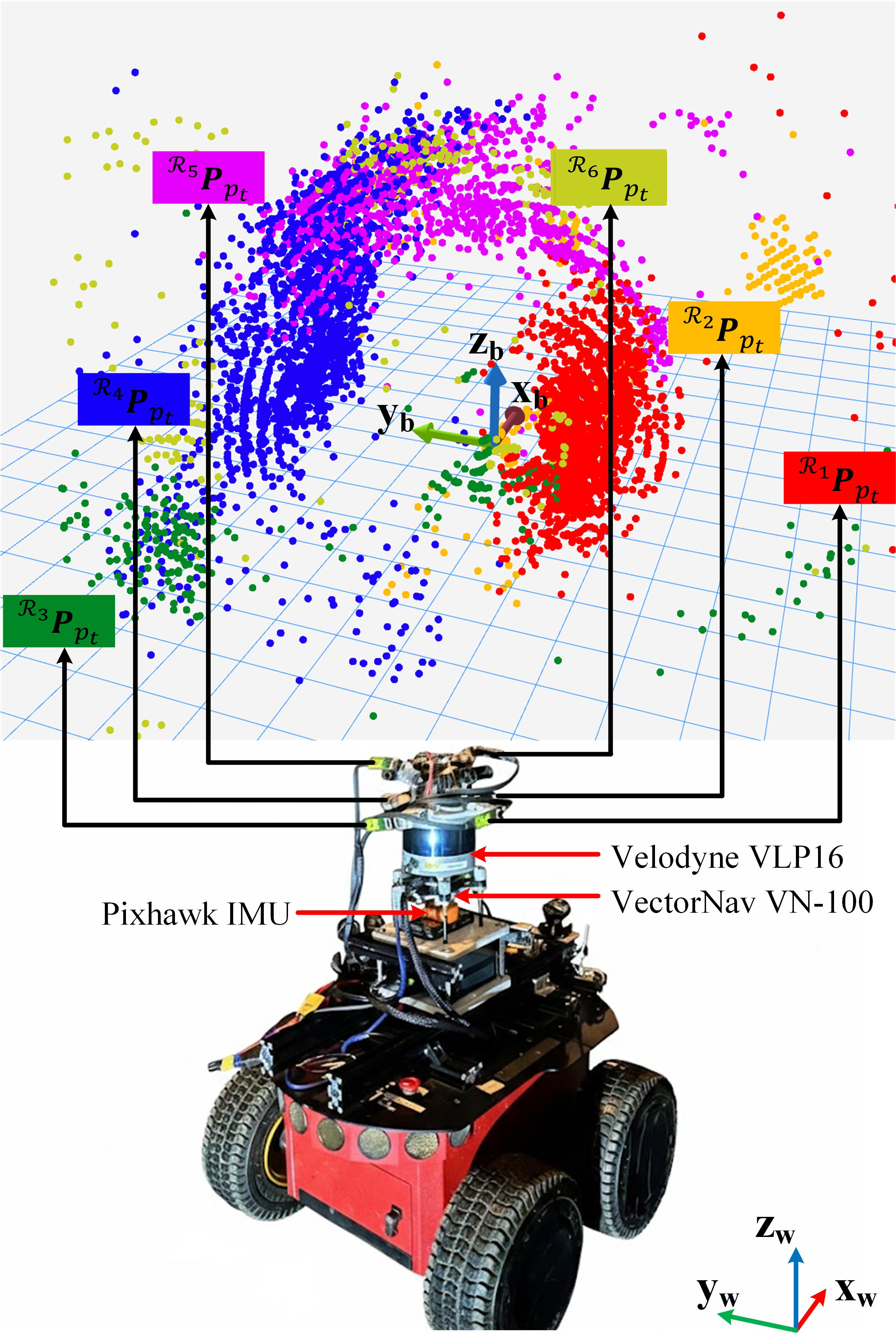}
\caption{Left: Experimental setup with six mmWave TI-IWR6843AoP radars ($R_1$–$R_6$), LiDAR, and IMU mounted on a Pioneer 3-AT2 platform. Radars $R_1$–$R_4$ are planar, while $R_5$–$R_6$ are tilted to capture ceiling observations. Right: Single-frame radar point cloud from the multi-radar setup, with colour-coded targets corresponding to individual radars.} 
\label{fig: sensor_configurations}
\end{figure}

Towards the development of the decentralized LRIO, this article considers a four-wheel Pioneer 3-AT2 robot equipped with multiple FMCW radars arranged for the coverage of $360^\circ$ perception. Specifically, four radars are each mounted on one side of the robot, spanning around the front, left, rear, and right at $90^\circ$ intervals to provide complete horizontal coverage. The remaining two radars are positioned at the front and rear and are tilted upward at $45^\circ$ angles to capture obstacles from the ceilings of the cave environment~(Fig.~\ref{fig: sensor_configurations}). Since radars work on the principle of the doppler effect, upon relative motion, they provide a set of target points within their field of view (FOV). In complex cave-like subterranean environments characterized by multiple branches and openings, relying on a single radar sensor may be insufficient for reliable navigation. As the robot moves through these environments, a radar facing an open space may produce sparse or unreliable measurements due to the absence of reflective surfaces. In such situations, the complementary radar configuration becomes essential for maintaining consistent and robust sensing performance.

On the other hand, a 3D Velodyne LiDAR is well known for high-accuracy localization and mapping. However, its performance degrades in the presence of smoke, fog and dust, which are typically unavoidable inside subterranean mining environments. To address these perception challenges, this article introduces a novel decentralized LiDAR and multi-radar inertial odometry framework~\cite{mukherjee2023fault} based on a robust, error-tolerant optimal information filter. The proposed decentralized multi-sensor fusion architecture supports selective integration of sensor modalities, allowing measurements from different radar configurations and LiDAR to be incorporated in a resilient manner. The design enables the odometry to gracefully eliminate temporal failures by selectively excluding one or more sensor modalities upon detecting any temporal anomaly. The overall methodology is presented in the following sections. First, an IMU drift-tolerant RIO~\cite{mukherjee2023fault} is applied, which serves as the backbone for the different RIO configurations. Next, the decentralized LiDAR radar inertial odometry (LRIO) framework is presented, which considers an adaptive threshold-based sensor-modality selection strategy.  
\section{Generic Algorithm for Robust Error-Tolerant Optimal Information Filter (RETOIF)}\label{sec: GARETOIF}
Typically, for subterranean navigation, the performance of LIO in the absence of smoke is expected to be superior, while RIO can be considered an auxiliary modality to support continuing navigation in specific conditions. To optimally integrate both LiDAR and radar modalities and obtain a resilient odometry, this article introduces a novel decentralized multi-sensor fusion approach. Towards this, a theoretical foundation for the decentralized fusion, supported by an adaptive threshold based selective augmentation of sensor modalities, is presented next. The proposed framework integrates $n$ heterogeneous sensors, each strategically addressing unique challenges within the operational environment. Moreover, the propagation of estimation theories within this framework necessitates control inputs from an inertial measurement unit (IMU), which, while crucial to the process, is treated as an external sensor not encompassed within the set of $n$ primary sensors. 

\begin{figure*}[h]
\centering
\includegraphics [width=1.1\textwidth]{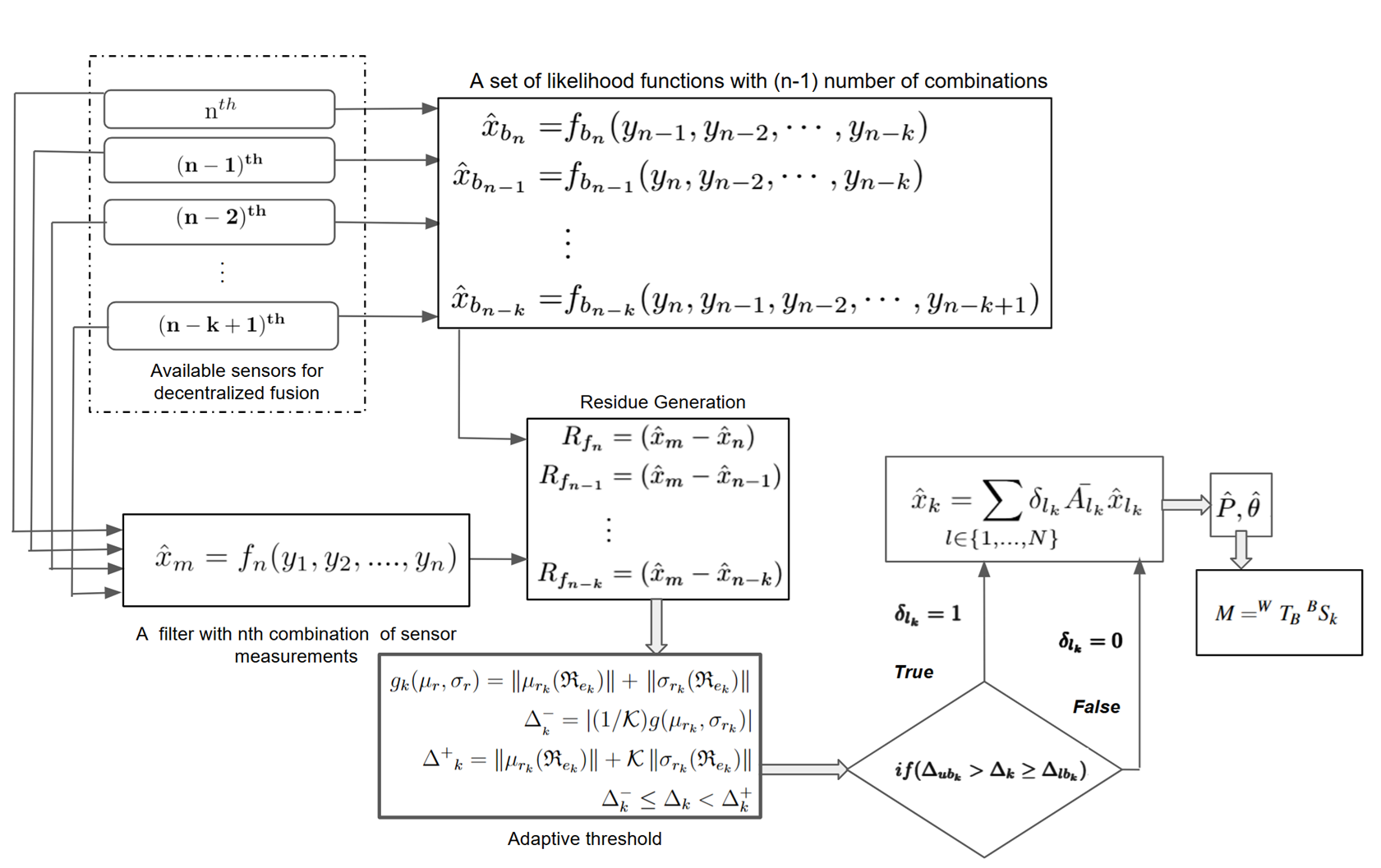}
\caption{\textbf{Generic Algorithm for the proposed decentralized Fusion~(RETOIF)}}
\label{fig: Concept_figure_entire_fusion_generic}
\end{figure*}

Following the acquisition of sensor measurements, a series of likelihood functions is constructed to create a filter bank, each incorporating (n-1) sensor measurements. Residual vectors are then generated by comparing the outputs of these individual filters with the outcomes of the likelihood function that fuses all $n$ sensor measurements into a single node. Once the residuals are computed, they are employed to synthesize the maximum likelihood estimator through adaptive threshold identification. In the final stage, the identified estimators are integrated using an optimal weighted function, ensuring precise sensor data fusion. Fig.~\ref{fig: Concept_figure_entire_fusion_generic} illustrates the generic concept figure of the proposed Robust Error-Tolerant Optimal Information Filter.
\subsection{Derivation of likelihood functions}\label{sec: likelihood_residues}
Let us consider a given dynamical system, with its mathematical model represented in state space form: $\mathscr{X}_{k+1}=f(\mathscr{X}_k), \mathscr{X}_k \in \mathbb{R}^{n_s}$ with initial condition $\mathscr{X}_{0}=\mathscr{X}(0)$. The output model of the system is considered as 
${z}_k=h(\mathscr{X}_k) \in \mathbb{R}^m$ 
For the above defined dynamical system, a set of measurements is obtained from multiple distinct sensor modalities or configurations, which serves as the basis for formulating the corresponding likelihood functions and is denoted by:
\begin{align}
{y} = \left[ {y}_1, {y}_2, \dots, {y}_n \right]\in \mathbb{R}^{m \times n},
\label{Eq: Equation y measurement}
\end{align}
where each ${y}_i \in \mathbb{R}^m$ for $i = 1, \dots, n$ denotes the measurement obtained from the $i^{th}$ sensor. The likelihood functions are formulated as individual state estimators, estimating the same state vector $\hat{\mathscr{X}}\in \mathbb{R}^{n_s}$ while considering $(n-1)$ different combinations of measured outputs from the combined output vector $y$ in the process of estimation. Hence, all such possible combination of output vectors constitutes a total number of $\binom{n}{n-1}$ likelihood functions expressed as:
\begin{align}\label{eq:cal_residue_vectors}
\hat{\mathscr{X}}_{b_n}&=f_{b_n}(y_{n-1},y_{n-2},\dots,y_{1})\\
\hat{\mathscr{X}}_{b_{n-1}}&=f_{b_{n-1}}(y_{n},y_{n-2},\dots,y_{1})\\ \nonumber
&\vdots \\
\hat{\mathscr{X}}_{b_{1}}&=f_{b_{1}}(y_{n},y_{n-1},\dots, y_3,y_2)
\end{align}
Note that, the indexed subscripts in each likelihood function $f_{b_K}: \mathbb{R}^{m \times (n-1)} \to \mathbb{R}^{n_s}$ 
for $ K={n, n-1, \dots, 1}$ indicate that the function is constructed using all combinations of the output vectors in $y$, except for the output vector corresponding to the indexed subscript, which is omitted. Additionally, each likelihood estimator is essentially an EKF. The collection of all these estimated states is combined and represented as $\hat{\mathcal{X}}_{b} = [\hat{\mathscr{X}}_{b_n}, \hat{\mathscr{X}}_{b_{n-1}}, \dots, \hat{\mathscr{X}}_{b_{1}}]\in \mathbb{R}^{n_s \times n}$.
Following a similar procedure, an additional likelihood function is constructed that utilizes all the available measurements from $y$, presented as
\begin{align}
\label{eq:estimators_n_sensors}
\hat{\mathscr{X}}_{n}&=f_{n}(y_1,y_2,....,y_{n}) 
\end{align}
This implies that the estimation process utilizes all $n$ sensor inputs to compute $\hat{x}_{n}$ with $f_n$ encapsulating a tightly coupled EKF formulation. This specific design choice of the likelihood function enables a residual generation mechanism to identify temporal anomalies in sensor measurements.
\subsection{Residue Generation}\label{sec: Residue_generation}
The residuals are formed by computing the difference between the reference state estimate $\hat{\mathcal{X}}_{n}$ denoting the centralized EKF where n represents all the available sensor modalities.
\begin{align}
\hat{\mathcal{X}}_{n}=\begin{bmatrix}
\hat{x}_{n_k} \\
\hat{y}_{n_k}
\end{bmatrix}\in \mathbb{R}^2
\end{align}
and the respective estimates from the bank of filters $\hat{\mathcal{X}}_{b_n},\hat{\mathcal{X}}_{b_{n-1}}~\dots~ \hat{\mathcal{X}}_{b_{n-K}}$. Individually, they can be written as 
\begin{align}
\mathcal{X}_{b_n}&=\begin{bmatrix} \hat{x}_{b_{n_k}} \\ \hat{y}_{b_{n_k}} \end{bmatrix}
\quad
\mathcal{X}_{b_{n-1}}=\begin{bmatrix} \hat{x}_{b_{n-1_k}} \\ \hat{y}_{b_{n-1_k}} \end{bmatrix}
\quad \dots \quad
\mathcal{X}_{b_{n-k}}=\begin{bmatrix} \hat{x}_{b_{n-k_k}} \\ \hat{y}_{b_{n-k_k}} \end{bmatrix}
\label{Eq: filter_states_first_layer}
\end{align}
Formally, the individual residuals are expressed as:
\begin{align}
\label{bank_est}
\mathfrak{R}_{f_n}&=(\hat{\mathcal{X}}_{n}-{\hat{\mathcal{X}}}_{0})\\
\mathfrak{R}_{f_{n-1}}&=(\hat{\mathcal{X}}_{n}-{\hat{\mathcal{X}}}_{n-1})\\ \vdots\\
\mathfrak{R}_{f_{n-K}}&=(\hat{\mathcal{X}}_{n}-{\hat{\mathcal{X}}}_{n-K})
\end{align}
where,  $\mathfrak{R}_{f_j} \in \mathbb{R}^{2}$ represents the residual associated with the $j$th filter and the indices $n,n-1,\cdots,n-K$ denote the respective combinations of sensor measurements utilized for state estimation. The residual vector $\mathfrak{R}_e$ is then constructed as:
\begin{align}
\label{eq:final_residue_com}
\mathfrak{R}_e=\begin{bmatrix}
\mathfrak{R}_{f_{n}} & \mathfrak{R}_{f_{{n}-1}} & \mathfrak{R}_{f_{{n}-2}} &\cdots  &\mathfrak{R}_{f_{{n}-K}} 
\end{bmatrix}^T\in \mathbb{R}^{K \times 2}
\end{align}
The stacked residual matrix $\mathfrak{R}_{e_k}$ plays a crucial role in characterizing the discrepancies between the state estimates generated by the centralized Kalman filter and the individual filters in the first-layer filter bank. Each residual vector $\mathfrak{R}_{f_i,k}$ represents the deviation resulting from the exclusion of a specific sensor measurement in the corresponding state estimation process. For the adaptive threshold based hypothesis decision, each residual vector is transformed into a scalar quantity using the Euclidean norm ($L_2$-norm). The scalar residual magnitude associated with the $i$th estimator at time step $k$ is defined as
\begin{equation}
\label{eq:scalar_residual}
r_{i,k} =
\left\|\mathfrak{R}_{f_i,k}\right\|_2.
\end{equation}
Here, $r_{i,k}\in\mathbb{R}{\geq 0}$ represents the magnitude of the residual vector and serves as the
scalar test statistic for comparison with the adaptive threshold bounds. These bounds are determined from the statistical properties of the residuals, as described in the subsequent section. Based on this comparison, estimators whose scalar residual magnitudes fall within the adaptive threshold region are retained as admissible candidates for the ultimate-layer fusion, whereas those outside the region are excluded. The resulting binary selection decisions determine which state estimates contribute to the final optimal fusion process.
\subsection{Adaptive Threshold Region}\label{sec: adaptive threshold}
As outlined in Section~\ref{sec: likelihood_residues}, residual vectors are computed for each configuration. Consequently, if $n$ likelihood functions are utilized, $n$ residual vectors are generated for each state at every time instance. Subsequently, these residual vectors are subjected to statistical analysis under the assumption that they all follow the normal distribution with non-zero mean.
\begin{align}\label{Eq: mean}   
\mu_{k}=\sum_{i=1}^{n}\frac{\mathfrak{R}_{e_ki}}{n}\\
\sigma_{k}=\sqrt{\frac{\sum_{i=1}^{n}(\mathfrak{R}_{e_ki}-\mu)^2}{n}} \label{Eq: variance}
\end{align}
\begin{figure*}[h]
\centering
\includegraphics [width=1.0\textwidth]{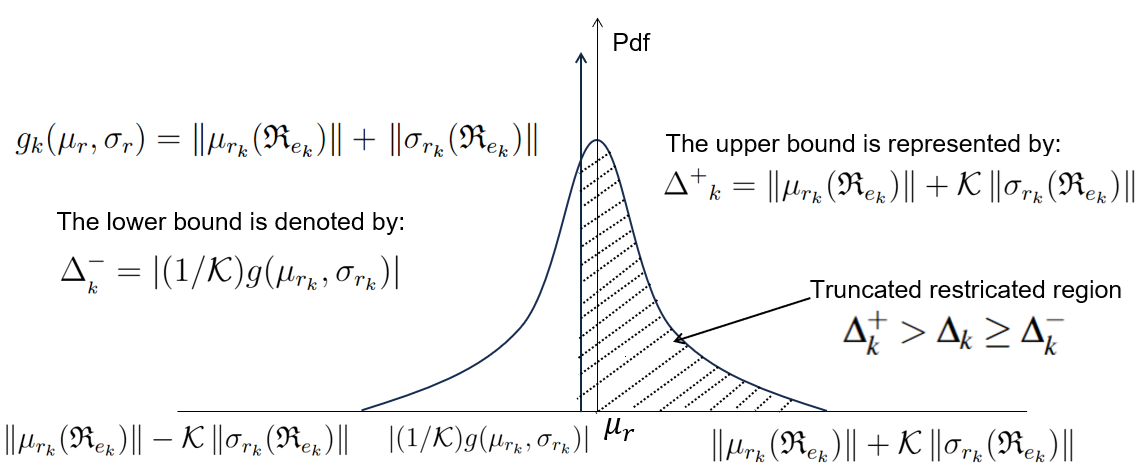}
\caption{The upper and lower bounds of the adaptive threshold are determined using the residual vectors, $\mathfrak{R}_{e_k}$, at each time instance. A statistical analysis is conducted on $n$ residuals under the assumption that they follow a normal distribution with nonzero mean. These bounds define the region of error for the maximum likelihood functions and serve as an accurate adaptive threshold for integrating pose estimates from multiple estimators in the final layer. The range of these threshold bounds can be adjusted using the tuning parameter $\mathcal{K}$, making the region dynamic and adaptive.}
\label{fig: Concept_Figure_Adaptive_Threshold}
\end{figure*}
To incorporate an adaptive mechanism for dynamically setting bounds, a truncated normal distribution is considered over this normal distribution region. To represent the lower and upper bounds, a function $g$ is introduced, illustrated in Fig.~\ref{fig: Concept_Figure_Adaptive_Threshold}, defined as the sum of the mean $\mu$~(Eq.~\ref{Eq: mean}) and the standard deviation $\sigma$~(Eq.~\ref{Eq: variance}). The upper bound $(\Delta^{+}_{k})$ is expressed as $(\mu(\mathfrak{R}_{e_k})+ K\sigma(\mathfrak{R}_{e_k}))$, where $K$ determines the dynamic and adaptive region. The lower bound $(\Delta^{-}_{k})$ is calculated as $\left \|(1/\mathcal{K})g(\mu_{r_k},\sigma_{r_k})\right \|$, accurately defining the variance region.
\begin{align} 
g_{k}(\mu_r,\sigma_r)=\left \|\mu_{r_k}(\mathfrak{R}_{e_k})\right \|+ \left \|\sigma_{r_k}(\mathfrak{R}_{e_k})\right\|\\
\Delta^{-}_{_k}=\left|(1/\mathcal{K})g(\mu_{r_k},\sigma_{r_k})\right|\\
\Delta^{+}{_k}=\left \|\mu_{r_k}(\mathfrak{R}_{e_k})\right\|+ \mathcal{K}\left \|\sigma_{r_k}(\mathfrak{R}_{e_k})\right \|\\
\Delta^{-}_k \leq \Delta_k < \Delta^{+}_k \label{Eq: adaptive weight update rule}
\end{align}
$\left | .\right |$ represents the modulus of variables. $\left \|.\right \|$ denotes a vector with a norm-2. Where $g$ is the function of $\mathfrak{R}_{e_k}$'s mean~$(\mu)$ and standard deviation~$(\sigma)$, and $\mathcal{K}\in \mathbb{R}^+$ is an adaptive gain that serves as a tuning parameter for adjusting the variance and is always a positive real number. 
\begin{equation}
    \Delta_k^- < \Delta_k^+
\end{equation}
because
\begin{align}
    \Delta_k^+ - \Delta_k^-
    &= \left(\mu_{r_k,k} +\mathcal{K}\sigma_{r_k,k}\right)
    - \frac{\mu_{r_k,k} + \sigma_{r_k,k}}{\mathcal{K}} \notag \\
    &= \frac{(\mathcal{K}-1)\left(\mu_{r_k,k} + (\mathcal{K}+1)\sigma_{r_k,k}\right)}{\mathcal{K}}.
\end{align}
For
\begin{equation}
    K > 1,  \qquad \mu_{r_k,k} > 0, \qquad \sigma_{r_k,k} > 0,
\end{equation}
we get
\begin{equation}
    \Delta_k^+ - \Delta_k^- > 0.
\end{equation}
For $\mathcal{K}>1$, the lower bound is always smaller than the upper bound. Therefore the interval between them is valid.
so, $\Delta_{k}^{-}<\Delta_{k}^{+}$, this valid region defined by the equation given below:  
\begin{equation}
    \Delta_k^- \leq \Delta_k<  \Delta_k^+
\end{equation}
Where $\Delta_k \in \mathbb{R}^n$ is the region of the maximum likelihood estimate at time step $k$. $\Delta^{-}_k \in \mathbb{R}^n$ be the lower bound vector at time step $k$. $\Delta^{+}_k \in \mathbb{R}^n$ be the upper bound vector at time step $k$ or it can be written as eq.~\ref{Eq: adaptive weight update rule},
\begin{align}
    \text{where} \quad 
    \Delta_k^- &= 
    \begin{bmatrix}
        \Delta_{k,1}^- \\
        \Delta_{k,2}^- \\
        \vdots \\
        \Delta_{k,n}^-
    \end{bmatrix}
    , \quad
    \Delta_k =
    \begin{bmatrix}
        \Delta_{k,1} \\
        \Delta_{k,2} \\
        \vdots \\
        \Delta_{k,n}
    \end{bmatrix}
    , \quad
    \Delta_k^+ =
    \begin{bmatrix}
        \Delta_{k,1}^+ \\
        \Delta_{k,2}^+ \\
        \vdots \\
        \Delta_{k,n}^+
    \end{bmatrix}
\end{align}
\subsection{Hypothesis test for adaptive threshold and Ultimate Layer for MLE Synthesis}\label{sec: Hypothesis test}
The estimators that fall within the adaptive region are considered Maximum Likelihood Estimators (MLEs), as outlined sec.~\ref{sec: adaptive threshold}. The adaptive thresholding mechanism defines an acceptance region for state estimators based on the statistical analysis of residual vectors. Let the adaptive threshold region at a time step $k$ be denoted as:
\begin{equation}
\Delta_k = \{r_{i,k} \mid \Delta^{-}_{k} \leq r_{i,k} \leq \Delta^{+}_k \}
\end{equation}
where, $r_{i,k}$ is the residual associated with the $i$~th estimator, $\Delta^{-}_k$ is the lower bound of the threshold, $\Delta^{+}_k$ is the upper bound of the threshold.
\begin{algorithm}[]
\caption{RETOIF}
\label{Algo:RETOIF-lrio-1}
\textcolor{black}{\begin{algorithmic}[1]
\Require $y \gets [y_1, y_2, \dots, y_n]^{\mathsf T}$
\Ensure $\hat{\mathscr{X}}_k$
\vspace{3pt}
\For{$i = 1 : n$}\\
  $\hat{\mathscr{X}}_{b_i} \gets f_b\!\big(\{y_j\}_{j \neq i}\big)$
\EndFor
\vspace{3pt}
\State $\hat{\mathscr{X}}_n \gets f(y_{1},y_{2}$,$\hdots$,$y_{n})$
\vspace{3pt}
\For{$i = 1 : n$}
    \State $\mathfrak{R}_{f_i} \gets \hat{\mathscr{X}}_n - \hat{\mathscr{X}}_{b_i}$
\EndFor
\vspace{3pt}
\State $\mu_k \gets \dfrac{1}{n} \sum_{i=1}^{n} \mathfrak{R}_{e_{k,i}}$
\State $\sigma_k \gets \sqrt{ \dfrac{1}{n} \sum_{i=1}^{n} (\mathfrak{R}_{e_{k,i}} - \mu_k)^2 }$
\vspace{3pt}
\State $\Delta_k^{-} \gets \left| \dfrac{1}{\mathcal{K}} \, g(\mu_{r_k}, \sigma_{r_k}) \right|$
\State $\Delta_k^{+} \gets \left\| \mu_{r_k}(\mathfrak{R}_{e_k}) \right\|+ \mathcal{K} \, \left\|\sigma_{r_k}(R_{e_k}) \right\|$
\vspace{3pt}
\For{$i = 1 : n$}
    \If{$r_{i,k} \in [\Delta_k^{-}, \Delta_k^{+}]$}
        \State $y_i \in \mathcal{Y}_{\text{fused}}$
    \Else
        \State $y_i \notin \mathcal{Y}_{\text{fused}}$
    \EndIf
\EndFor
\vspace{3pt}
\State $W_k \gets \Sigma_k^{-1} e_{\delta_k} ( e_{\delta_k}^T \Sigma_k^{-1} e_{\delta_k} )^{-1}$
\State $\hat{\mathscr{X}}_k \gets \sum_{l=1}^{n} \delta_{l_k} \bar{A}_{l_k} \hat{x}_{l_k}$
\State $\mathbb{P}_k \gets ( e_{\delta_k}^T \Sigma_k^{-1} e_{\delta_k} )^{-1}$
\State \Return $(\hat{x}_k, \mathbb{P}_k)$
\end{algorithmic}}
\end{algorithm}
 The hypothesis decision rule becomes:
\begin{align}
\text{Null hypothesis}: H_0 :r_{i,k}\notin [\Delta^{-}_{k},\Delta^{+}_{k}]\\
\text{Alternative hypothesis}: H_1 :r_{i,k}\in [\Delta^{-}_{k},\Delta^{+}_{k}]
\end{align}
Based on the region $\Delta_k$, the $\delta_{lk}$ takes a value of either 0 or 1, which can be mathematically expressed as:
\begin{align}
\label{eq:logical_operation}
if (\Delta_{ub_k}>r_{i,k}\geq \Delta_{lb_k}),~~~~\delta_{lk}=1\\
   else, \delta_{lk}=0,
\end{align}
$\delta_{l_k}\in\{0,1\}$ where~$\delta_{l_k}$ represents the hypothetical output from the autonomous threshold bound. The final layer combines the accurate pose estimates from this comparison to produce the ultimate fused pose estimation.
\begin{align}
    \hat{x}_{k}=\sum_{\mathclap{l \in \{1,\dots, n\}}}\delta_{l_k}\bar{A_{l_k}}\hat{x}_{l_k}
\end{align}
Where $\bar{A}_{l_k}$ represents the weight associated with the filters from the first layer, $\hat{x}_{l_k}$ represents the pose estimates derived from the likelihood functions, and $\delta_{lk}$ is the logical output resulting from the comparison. The covariance~$P_k$ is defined as:
\begin{align}
P_k=E(\hat{x}_k\hat{x}^T_k)=W^T_k\Sigma_{k}W_k
\end{align}
where $W_k$ is the weight matrix and $\Sigma_k=diag(\Sigma_{1_k},\Sigma_{2_k},\dots,\Sigma_{n_k})$ is the block diagonal covariance matrix of individual estimates. The objective is to minimize the trace of the covariance matrix:
\begin{align} \label{eq:adance_oif}
  &\min_{\Bar{A}_{1_k},\dots, \Bar{A}_{n_k}} ~J_k=\frac{1}{2}tr(P_k)=\frac{1}{2}tr(W_k^T\Sigma_kW_k)\\
&\textit{subjected to} ~\left(\bold{W}_k^T\boldsymbol{e_{\delta_k}} -I \right)=\bold{0}
\end{align}
The weight matrix $W_k$ is defined as:
\begin{align}\label{Eq: Weight_matrix}
W_k=[\Bar{A}_{1k},\Bar{A}_{1k},~\dots~,\Bar{A}_{nk}]\\
e_{\delta_k}=[\delta_{A_k}I,\delta_{B_k},\dots,\delta_{G_k}I]^T
\end{align} 
$\Sigma$ is the state covariance matrix and $W_k$ is a matrix containing the weights. where, $\boldsymbol{e_{{\delta}_k}}=\left[\delta_{A_k}I,~ \delta_{B_k}I,~ \dots, ~\delta_{G_k}I \right]^T$. To solve the optimisation problem with the equality constraint, the Lagrange multiplier method is used. The augmented cost function is:
\begin{equation}
    \bar J_k=\frac{1}{2}tr(\bold{W}_k^{T}\Sigma_k \bold{W}_k)+tr\left[\Lambda_k\left( \bold{W}_k^T\boldsymbol{e_{{\delta}_k}} -I \right)\right]
\end{equation}
where, $\Lambda_k \in R^{n \times n}$ is the Lagrange multiplier. Evaluating the necessary conditions of optimality, $\frac{\partial \bar J_k}{\partial \bold{W}_k} = 0$ and $\frac{\partial \bar J_k}{\partial \\\Lambda_k}= 0$, leads to the following matrix equation:
\begin{equation} \label{eq:optimization sol}
 \begin{bmatrix}
\Sigma_k &\boldsymbol{e_{{\delta}_k}}\\ 
 \boldsymbol{e_{{\delta}_k}}^T& \bold{0}
\end{bmatrix} \begin{bmatrix}
\bold{W}_k\\ 
\Lambda_k
\end{bmatrix} = \begin{bmatrix}
\bold{0}\\ 
I
\end{bmatrix}
\end{equation}
The optimal weight matrix $\bm{W}_k$ is obtained as:
\begin{equation}
\bold{W}_k =
{\Sigma}_k^{-1}\boldsymbol{e_{{\delta}_k}}(\boldsymbol{e_{{\delta}_k}}^{T}{\Sigma}_k^{-1}\boldsymbol{e_{{\delta}_k}})^{-1}
\end{equation}
This result provides the optimal weights for fusing pose estimates, ensuring that the covariance trace is minimized under the given constraints. Using the expression for $W_k$, the covariance $P_k$ becomes:
\begin{align}\label{Eq: advanced_cov}
P_k = 
\left( \Sigma_k^{-1} \boldsymbol{e}_{\delta_k} 
\left( \boldsymbol{e}_{\delta_k}^T \Sigma_k^{-1} \boldsymbol{e}_{\delta_k} \right)^{-1} 
\right)^T 
\Sigma_k 
\left( \Sigma_k^{-1} \boldsymbol{e}_{\delta_k} 
\left( \boldsymbol{e}_{\delta_k}^T \Sigma_k^{-1} \boldsymbol{e}_{\delta_k} \right)^{-1} 
\right)
\end{align}

Taking the transpose of the first term in Eq.~\ref{Eq: advanced_cov}
\begin{align}\label{eq: T_cov_term}
\left( \Sigma_k^{-1} \boldsymbol{e}_{\delta_k} 
\left( \boldsymbol{e}_{\delta_k}^T \Sigma_k^{-1} \boldsymbol{e}_{\delta_k} \right)^{-1} 
\right)^T
=\left( \left( \boldsymbol{e}_{\delta_k}^T \Sigma_k^{-1} \right) 
\left( \boldsymbol{e}_{\delta_k}^T \Sigma_k^{-1} \boldsymbol{e}_{\delta_k} \right)^{-1} 
\right)
\end{align}
and substitute Eq.~\ref{eq: T_cov_term} into the equation~\ref{Eq: advanced_cov}, the final expression for the covariance is obtained:
\begin{align}
P_k =&\left( \boldsymbol{e}_{\delta_k}^T \Sigma_k^{-1} \boldsymbol{e}_{\delta_k} \right)^{-1}\label{Eq: Final_expression}
\end{align}
This formulation ensures that the MLE selection is statistically robust, resilient to noise, and effectively excludes outliers. This study introduces a comprehensive multi-sensor fusion framework that integrates multiple radars, LiDAR, and an inertial measurement unit (IMU) to achieve robust and accurate pose estimation. The framework evaluates different radar configurations and their corresponding RIOs, yielding four distinct RIO outputs, each associated with a specific field of view (FOV) to address obstacle dependent challenges in the environment. In addition, the LIO SLAM module is incorporated to provide an additional independent source of pose information. Together, these components produce a total of five pose estimates, offering a versatile and redundant fusion scheme that enables reliable localization under diverse and challenging operational conditions.
\begin{figure}[h]
\centering
\includegraphics [width=1.0\textwidth]{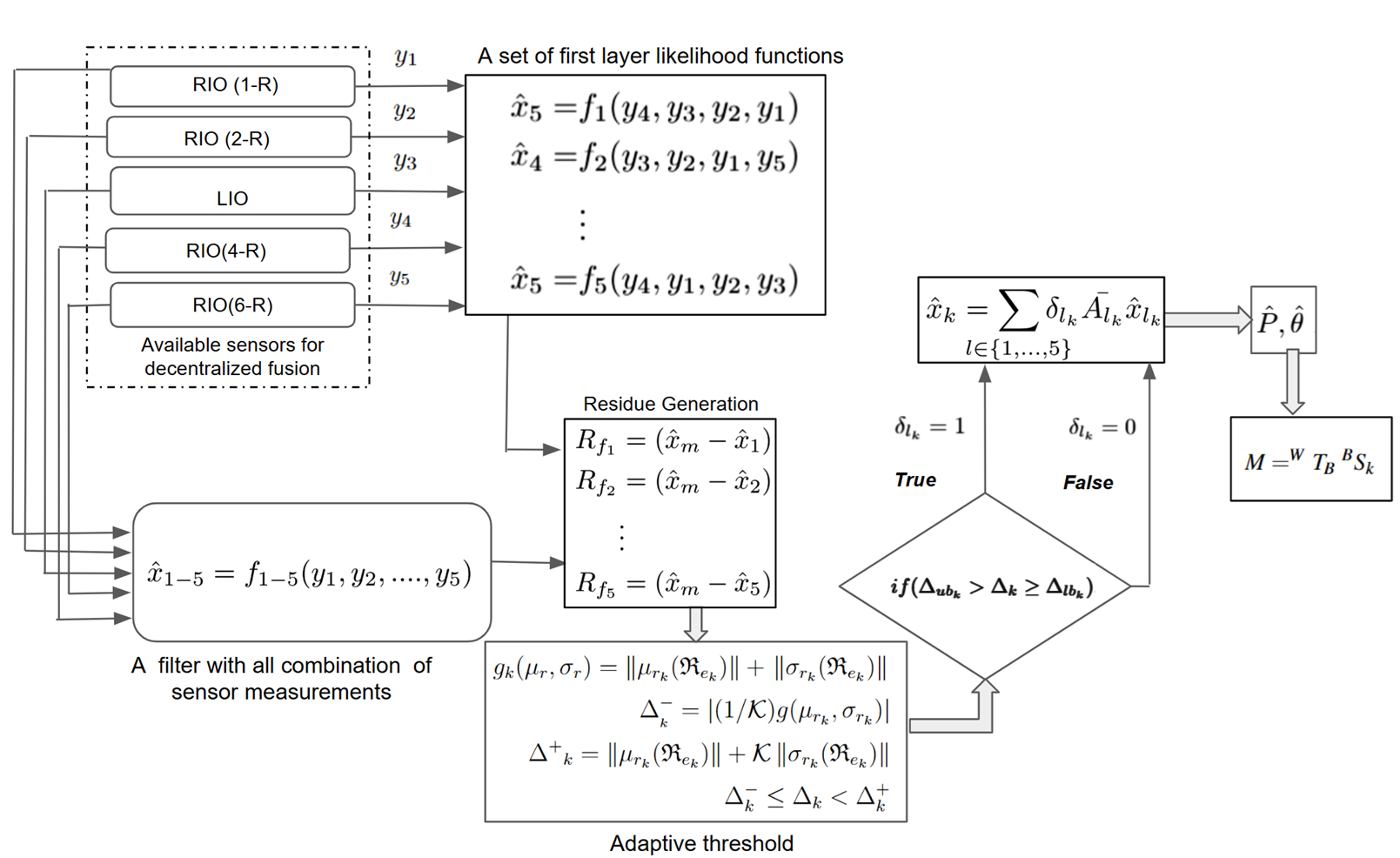}
\caption{RETOIF is designed for LiDAR Radar Inertial Odometry (LRIO). In this framework, RIO~(1-R) denotes a single front-radar setup, RIO(2-R) uses front and rear radars, RIO~(4-R) incorporates radars placed at the front, rear, left, and right, and RIO~(6-R) uses a six-radar configuration.}
\label{fig: Concept_figure_LRIO}
\end{figure}
Centralized sensor fusion is the integration of complementary sensors, such as LiDAR and radar, which is essential for achieving robust and accurate perception in challenging environments. LIO SLAM provides high-profile pose and map information, while radar offers reliable measurements in adverse conditions such as dense smoke, fog, or rain. This study explores a centralized sensor fusion framework~\cite{mukherjee2022resilient} that fused together all information in an Extended Kalman Filter(EKF), aiming to get localization accuracy and environmental understanding in GPS-denied environments.
\begin{figure*}[h]
\centering
\begin{subfigure}{0.48\textwidth}
\includegraphics[width=1.0\textwidth]{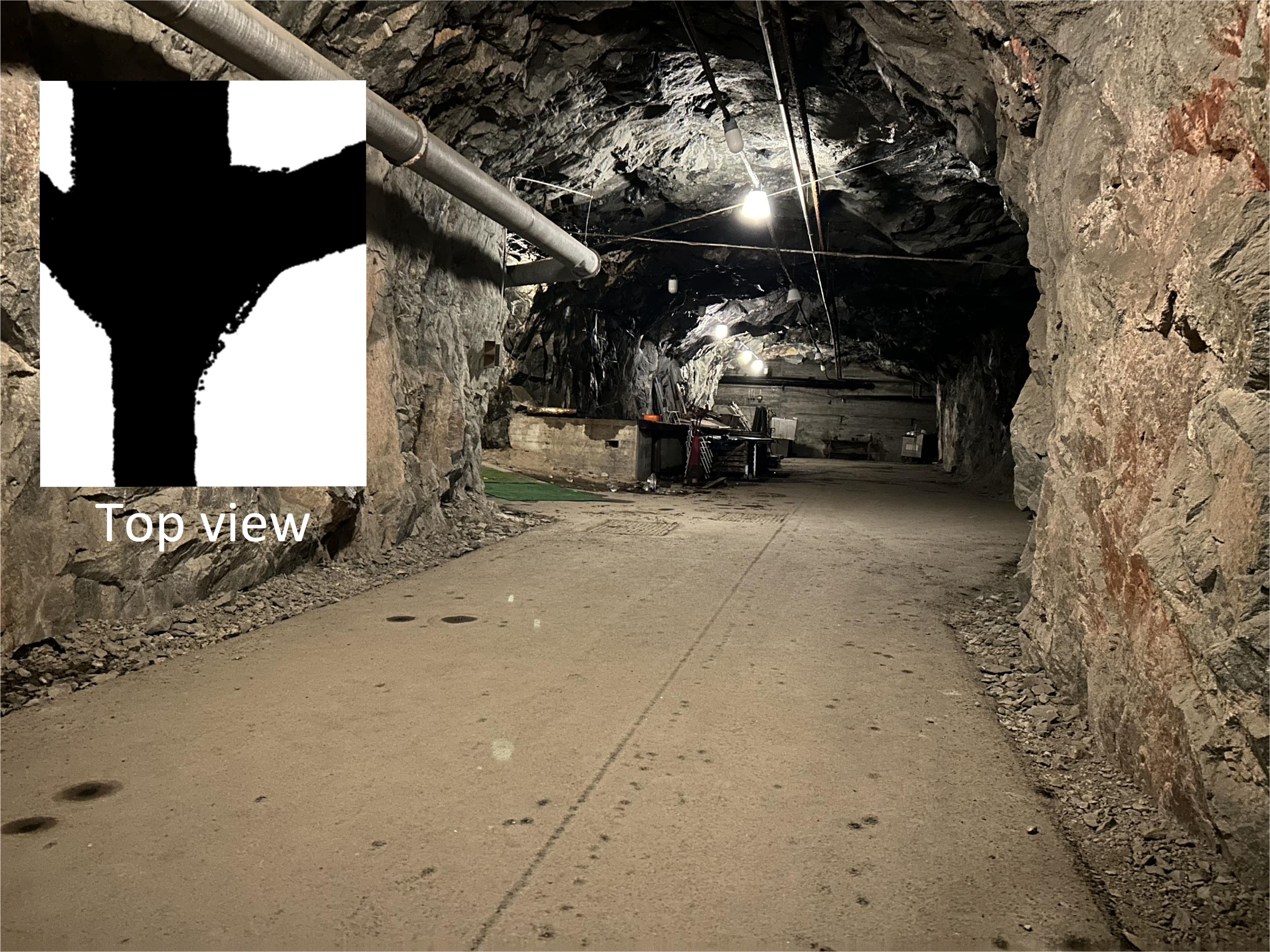}
\caption{\centering \textbf{Environment-1}: Experimental environment without smoke }
\label{fig:Environment_1}
\end{subfigure}
\begin{subfigure}{0.48\textwidth}
\includegraphics[width=1.0\textwidth]{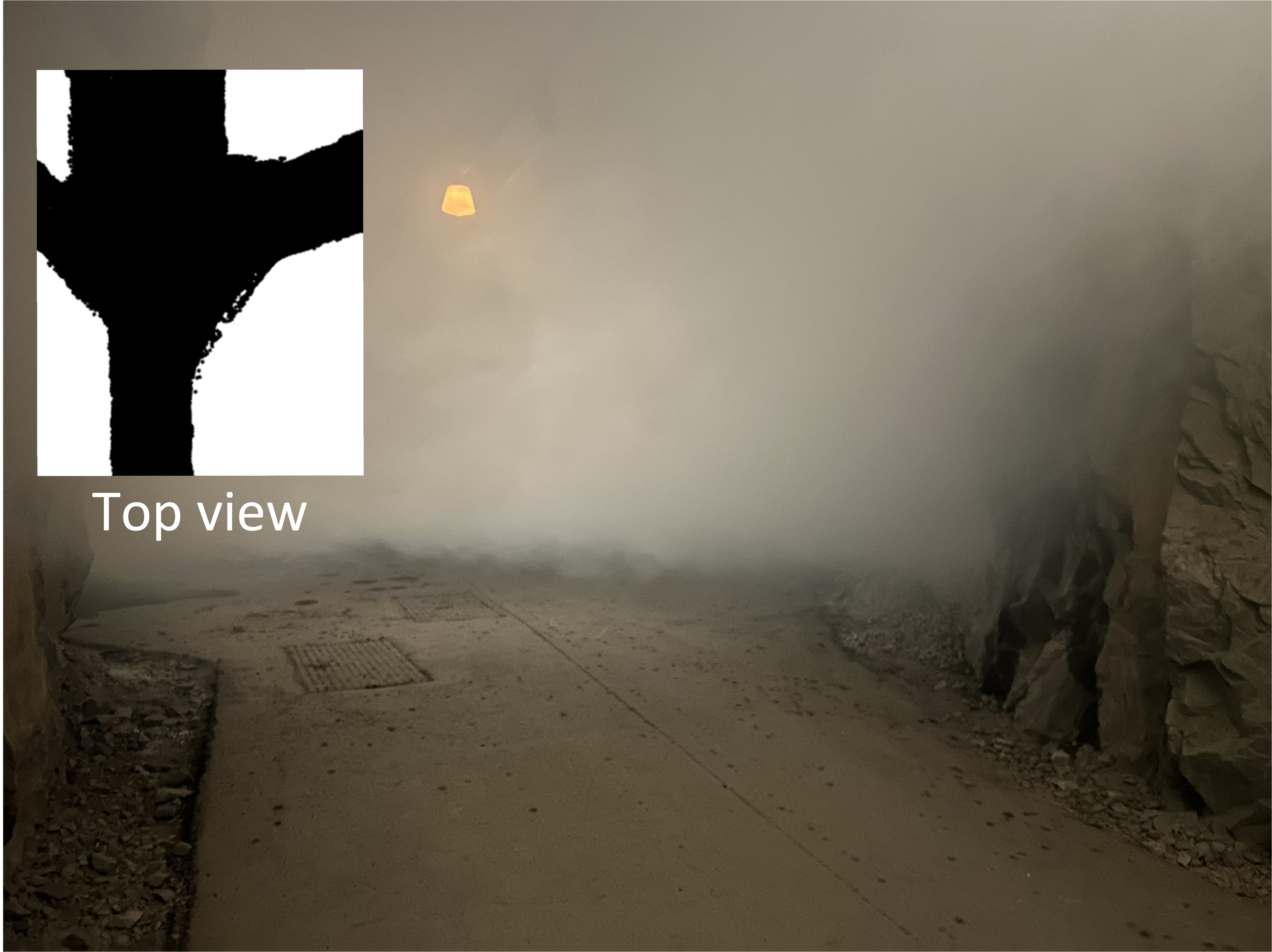} 
\caption{\centering \textbf{Environment-2}: Experimental environment in presence of smoke}
\label{fig:Environment_2}
\end{subfigure} 
\caption{Representative experimental environment inside Sub-T tunnel with and without smoke. A map of the experimental area is shown in the upper left corner of each image and was generated from the 3D LiDAR measurements.}
\label{fig:Exp_env_with_and_without_smoke}
\end{figure*}
The proposed decentralized  RETOIF is a novel approach designed to enhance the resilience and accuracy of the LRIO framework. Since we have considered four distinct radar configurations and LIO that result in five redundant pose measurements, these are utilized to construct the likelihood filter bank. Each filter within the bank combines sensor measurements using an EKF with the exception of one sensor from the entire sensor setup, as outlined in Section~\ref{sec: GARETOIF}.
\begin{table}[h]
\footnotesize
\centering
\caption{Configuration Parameters for 3D-Velodyne LiDAR}
\setlength{\tabcolsep}{2.25mm}
\renewcommand{\arraystretch}{1.2} %
\begin{tabular}{|c|c|c|c|c|}
\hline
\centering
\textbf{Parameter} & \textbf{Value}\\ \hline
\centering
\textbf{Laser Channels} &16  \\ \hline
\textbf{Vertical Resolution} &0.2° \\ \hline
\textbf{Field of View (FOV)} & 360°  \\ \hline
\textbf{Range} & 100m  \\ \hline
\textbf{Rotation Rate} &(1-100) Hz\\ \hline
\textbf{Points per second} &~300,000 points/sec  \\ \hline
\end{tabular}
\label{Table: 3D velodyne Lidar}
\end{table}
For example, the $5^{th}$ filter in the bank is designed to fuse all sensor measurements, except those measurements coming from the sensor setup assigned to the fifth sensor from the total sensor configuration. The $\hat{\mathcal{X}}_{b_k}$ estimated states contribute to the yield of residue vectors, which are then compared with the estimated states from an EKF that integrates the outcomes of the complete sensor setup into a single node, as defined by Eqs.~\ref{eq:cal_residue_vectors}-\ref{eq:estimators_n_sensors}. After deriving the residual vectors from the likelihood functions, we conduct a statistical analysis using their mean and standard deviation, as detailed in Sec.~\ref{sec: adaptive threshold}. We assume that the residual vectors for each state, denoted as $\mathfrak{R}_{f_1}$, $\mathfrak{R}_{f_2}$, $\mathfrak{R}_{f_3}$ and $\mathfrak{R}_{f_5}$ follow a normal distribution with means~$\mu(\mathfrak{R}_{f_k})$, where these means are non-zero. This thresholding algorithm determines whether an error vector belongs to a specific statistical region. According to the hypothesis, the region populated with residuals is treated as the maximum likelihood estimators for every sampling time. The rest of the errors are considered to be outliers, the estimated states responsible for these errors are automatically excluded through the isolation framework. 
\begin{table}[h]
\footnotesize
\centering
\caption{Configuration Parameters for Millimeter-wave TI-IWR6843AoP Radar}
\setlength{\tabcolsep}{2.25mm}
\renewcommand{\arraystretch}{1.2} 
\begin{tabular}{|c|c|c|c|c|}
\hline
\centering
\textbf{Parameter} & \textbf{Value}\\ \hline
\centering
\textbf{Frame-rate} & 8.64-10 GHz\\ \hline
\textbf{Band width} & 60-64Hz \\ \hline
\textbf{Maximum range} & 8-10 meters   \\ \hline
\textbf{Maximum doppler} &  8-10m/s \\ \hline
\textbf{Range resolution} &0.047  \\ \hline
\textbf{Doppler velocity resolution} &3 cm/s   \\ \hline
\end{tabular}
\label{Table: radar_specifications}
\end{table}
This flexibility allows the thresholding mechanism to adapt to various sensor measurement scenarios. 

\section{Experimental Evaluations} 
The experimental data were collected using a Pioneer 3-AT2 platform in an underground area at the subterranean Lab of Lule\aa{} University of Technology (LTU) located in Lule\aa{}, Sweden~\cite{koval2022dataset}. Specifically, the experiments took place in a junction area of the underground tunnel, as illustrated in Fig~\ref{fig:Exp_env_with_and_without_smoke}. The robot was deployed in a subterranean tunnel under two environmental conditions to evaluate odometry accuracy in an infrastructure-free, GPS-denied SubT environment. Experiments were conducted in both smoke-free and smoke-filled scenarios. The smoke-free condition was used for quantitative comparison with a 3D Velodyne LiDAR (Table~\ref{Table: 3D velodyne Lidar}), as the tunnel lacks external localization infrastructure such as GPS or a motion capture system~(e.g., Vicon). Consequently, LiDAR SLAM is adopted as the ground truth, given its higher reliability under clear air conditions. The first environmental scenario, referred to as \textbf{Environment 1}, is shown in Fig.~\ref{fig:Environment_1}, while the second scenario, \textbf{Environment 2}, is depicted in Fig.~\ref{fig:Environment_2}. In Environment-1, the performance of the MRIO is evaluated under multiple sensor configurations using IMU and FMCW IWR6843AoP ES2.0~\cite{kyuroson2023multimodal} radars~(\ref{Table: radar_specifications}), specifically, we consider RIO configurations with one, two, four and six radars denoted as RIO~(1-R), RIO~(2-R), RIO~(4-R), and RIO~(6-R) respectively. 1-R is mounted at the front, 2-R is mounted at the front and rear, 4-R is mounted at the front, rear, left, and right, and 6-R is mounted at the front, rear, left, right, upper tilted front, and upper tilted right positions on the robot. Since each sensor has inherent limitations, the entire experimental path was measured using a measuring tape before starting the experiment along both straight~($11$ m) and curved segments~($15$ m) to evaluate the accuracy and effectiveness of the individual approaches.
\begin{figure*}[h]
\centering
\includegraphics[width=1.0\textwidth]{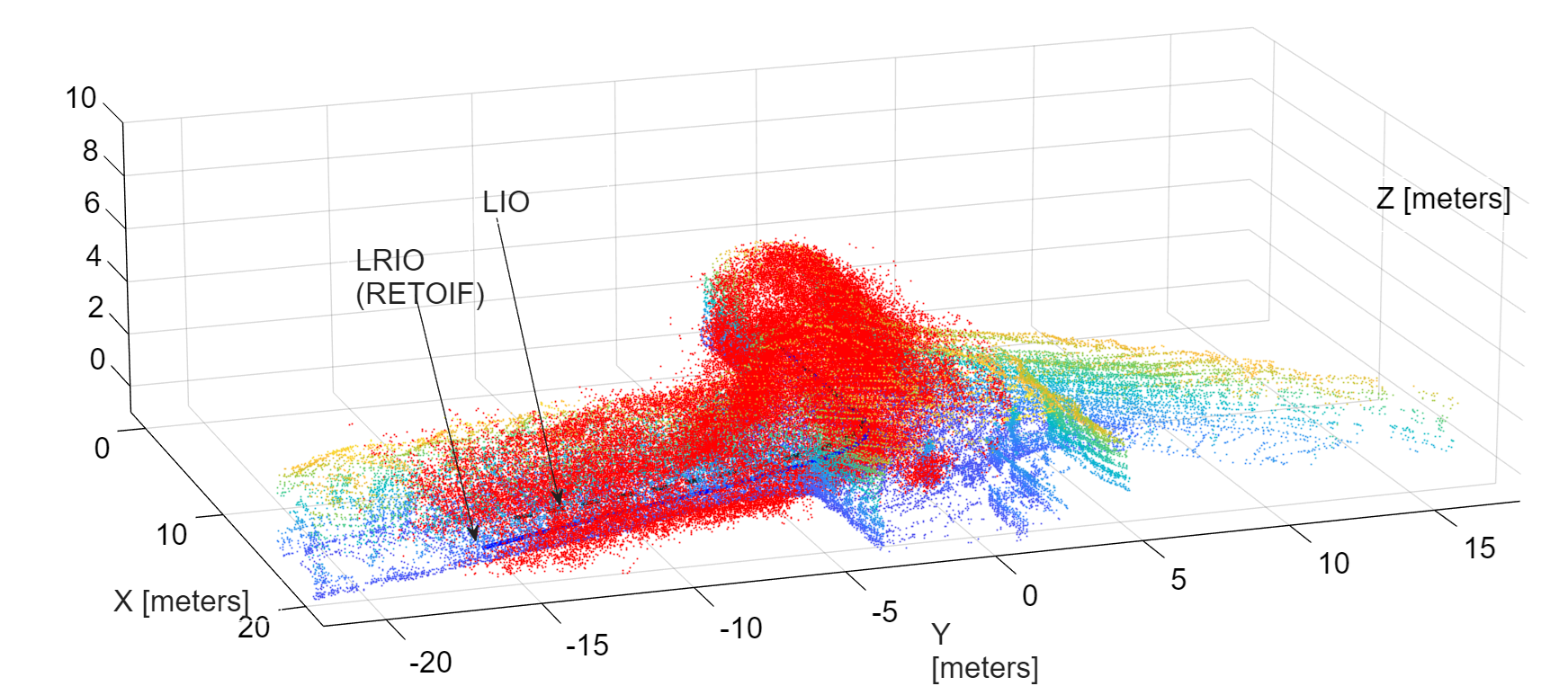}
\caption{The LRIO-based localization was compared with LIO. The radar body point clouds were transformed based on the pose estimates from the proposed decentralized framework. While the LiDAR can provide reliable measurements in the absence of smoke, its inherent sensor limitations result in less accurate odometry estimates compared to measurements obtained using a measuring tape.}
\label{fig:LRIO_based_lcal_radar_map_compared_LIO}
\end{figure*}
\begin{figure*}[h]
\centering
    \includegraphics[width=0.99\linewidth]{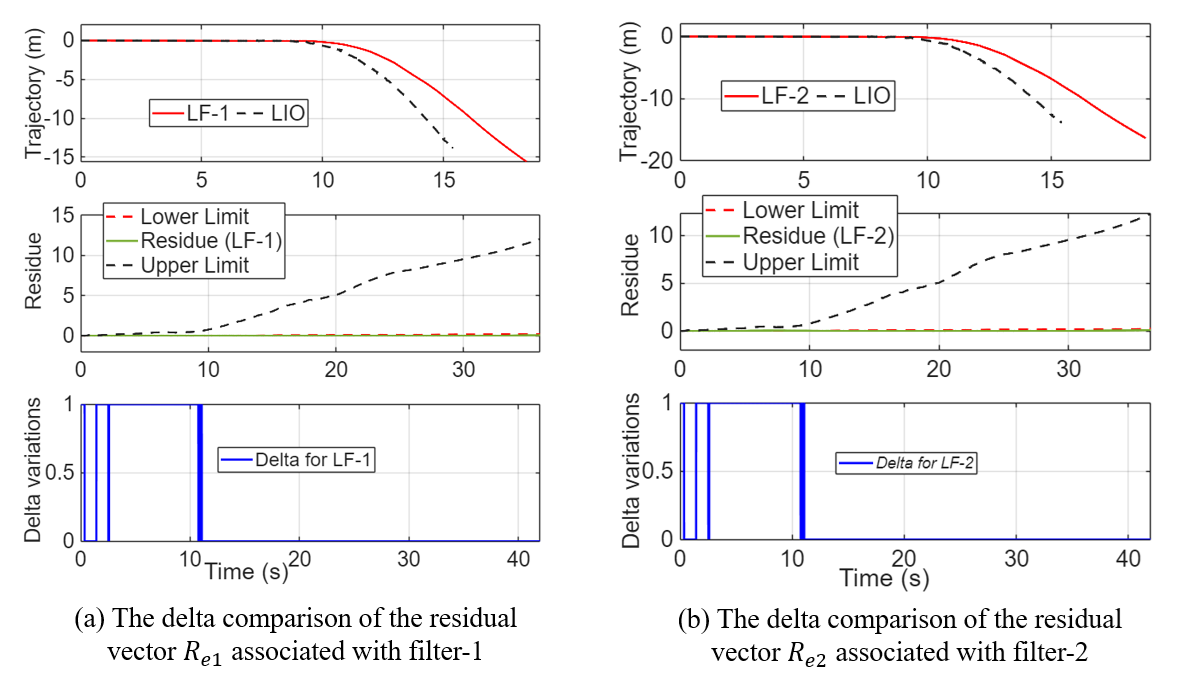}
\caption{The assessment of the first-layer likelihood functions (filters 1 and 2), incorporating the lower and upper threshold bounds, conducted in a smoke-free subterranean environment.}
\label{fig: Nosmk_filter_1_2}
\end{figure*}
\begin{figure*}[h]
\centering
\includegraphics[width=0.99\linewidth]{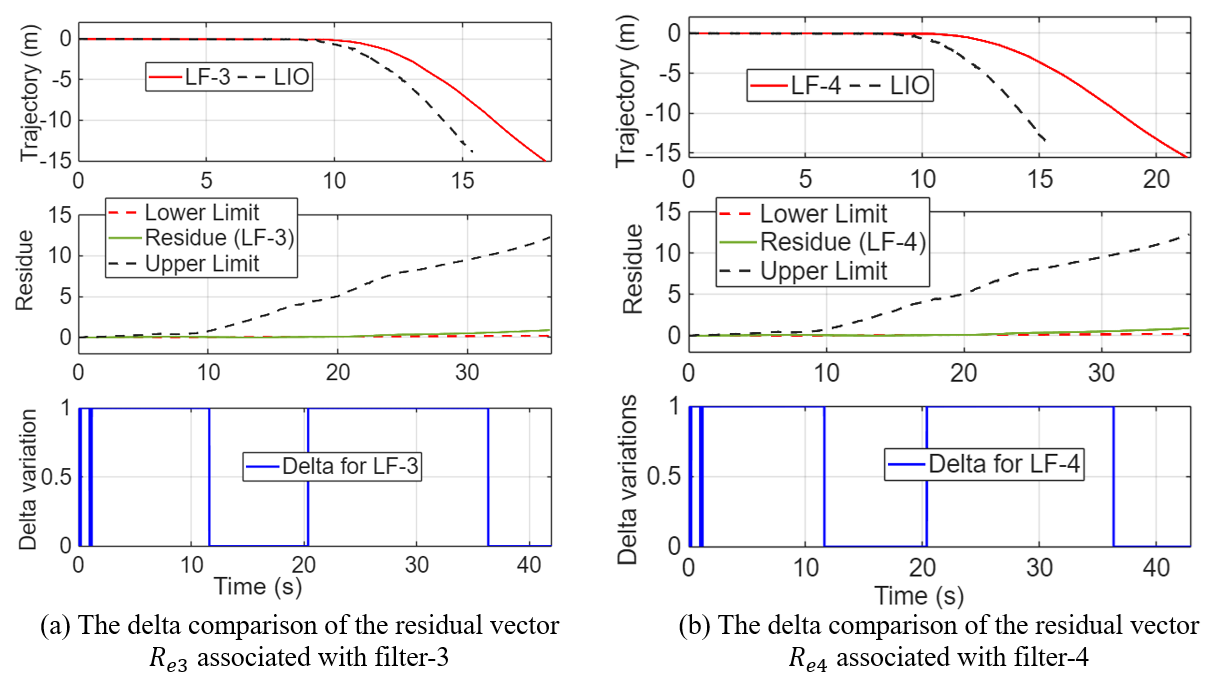}
\caption{The assessment of the first-layer likelihood functions (filters 3 and 4), incorporating the lower and upper threshold bounds, conducted in a smoke-free subterranean environment.}
\label{fig: Nosmk_filter_3_4}
\end{figure*}
LiDAR sensors are susceptible to smoke, making them unsuitable for comparison validation in smoke-filled environments (Environment-2). As a result, no sensor can reliably serve for performance comparison under smoke-filled conditions, leaving us to depend solely on measurements taken with a measuring tape as the ground truth for all the scenarios. During the experiments, the radars collected detailed data as the robot navigated the tunnel. These included radar target points~($x$,$y$,$z$) and corresponding Doppler velocities~($v_d$) under various conditions, which can significantly generate radar inertial odometry(RIO). Since subterranean environments are characterised by extreme cold and uneven conditions, and have a mild, gradually inclined tunnel, there is a higher possibility of the same phenomenon as in the article~\cite{mukherjee2026resilient} regarding the dynamical IMU acceleration bias, which a standalone EKF cannot address, since other methods have no certainty of working in such environmental adversities. This MRIO is considered for localization in all radar configurations. 

The placement and configuration of the radars are critical for localization and mapping performance, as they directly affect ego-velocity estimation. Performance discrepancies arise from variations in the radar field of view (FOV) and the presence of obstacles within the sensed region. Navigating in an unfamiliar and unstructured environment makes it extremely difficult to identify the most dependable sensor setup for high-accuracy localization, as LIO SLAM performance significantly deteriorates in dense fog, and no single radar configuration consistently provides the best results in all environmental scenarios. Thus, multi-sensor fusion concepts are needed to address these discrepancies coming from these distinct sensors and setups. In order to incorporate a multi-sensor fusion framework, we propose a novel decentralized algorithm that integrates an outlier detection mechanism based on a tightly coupled EKF-based filter bank~\cite{mukherjee2023fault}, along with a novel adaptive residual thresholding approach complemented by an outlier isolation framework~\cite{mukherjee2021decentralized} for effective outlier detection and isolation. This integrated design enhances both estimation accuracy and resilience across diverse operating environments. To evaluate the performance of the proposed architecture, three distinct datasets are considered: one collected in a smoke-free environment and two captured in smoke-filled conditions~(illustrated in Fig.~\ref{fig:In_smoke_scans_pointclouds_LiDAR_radar}), all recorded data are over identical distances and alignments for consistent evaluation. Within these smoke-filled environments, one is with highly dense smoke and labelled as Case-1, and another is with moderately dense smoke labelled as Case-2 for clarity and reference.
\begin{figure*}[h]
\centering
 \includegraphics[width=0.99\linewidth]{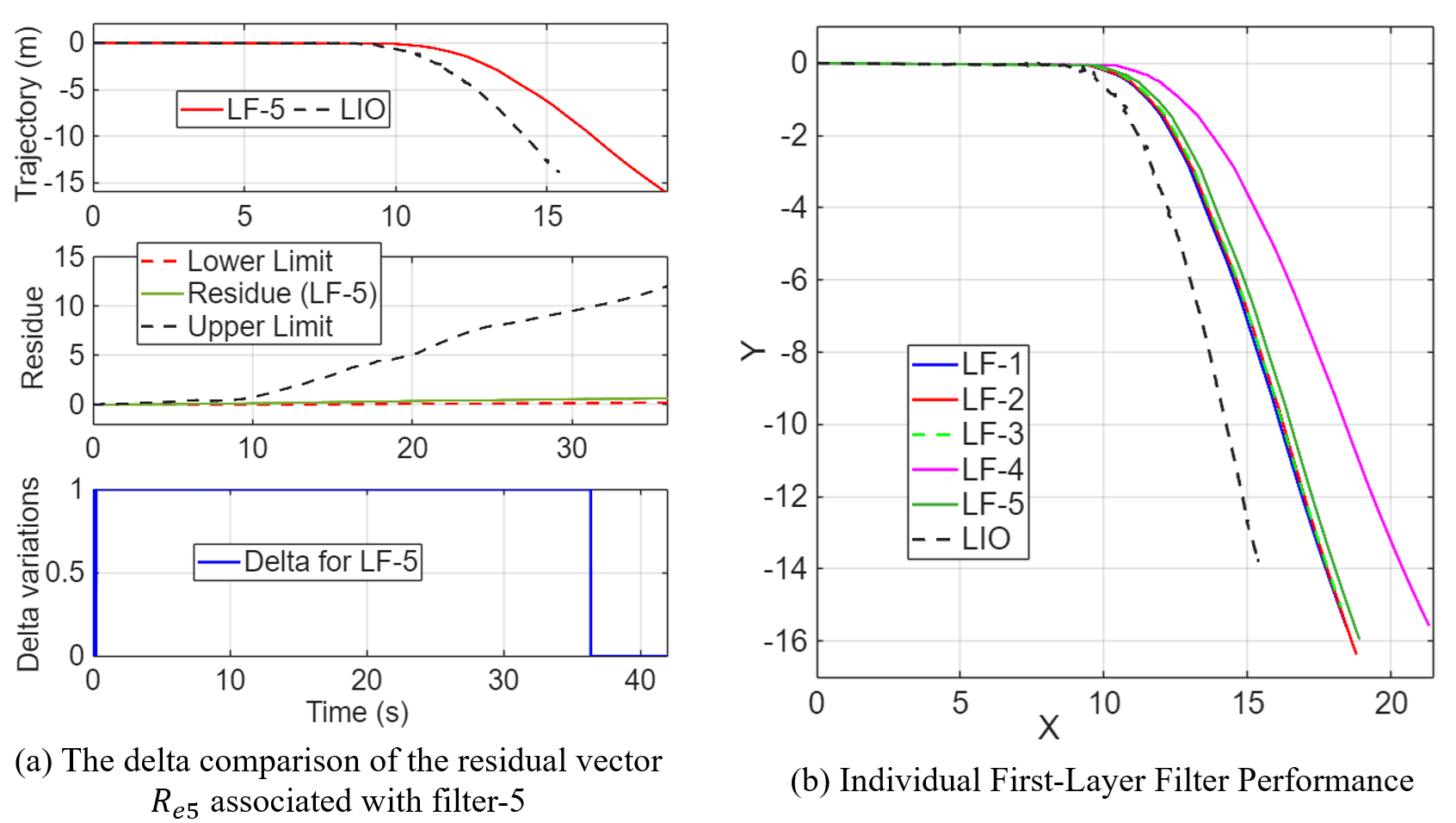}
\label{fig:All_filters_no_smoke}
\caption{The assessment of the first-layer likelihood function, with lower and upper bounds of residual thresholding, where Figure (b) represents estimated trajectories from all the filters compared with LIO}
\label{fig: Nosmk_filter5_all_odom}
\end{figure*}
\subsection{Experimental Results for Environment-1}\label{sec: Env1_results}
\begin{figure*}[h]
    \centering
\includegraphics[width=0.99\linewidth]{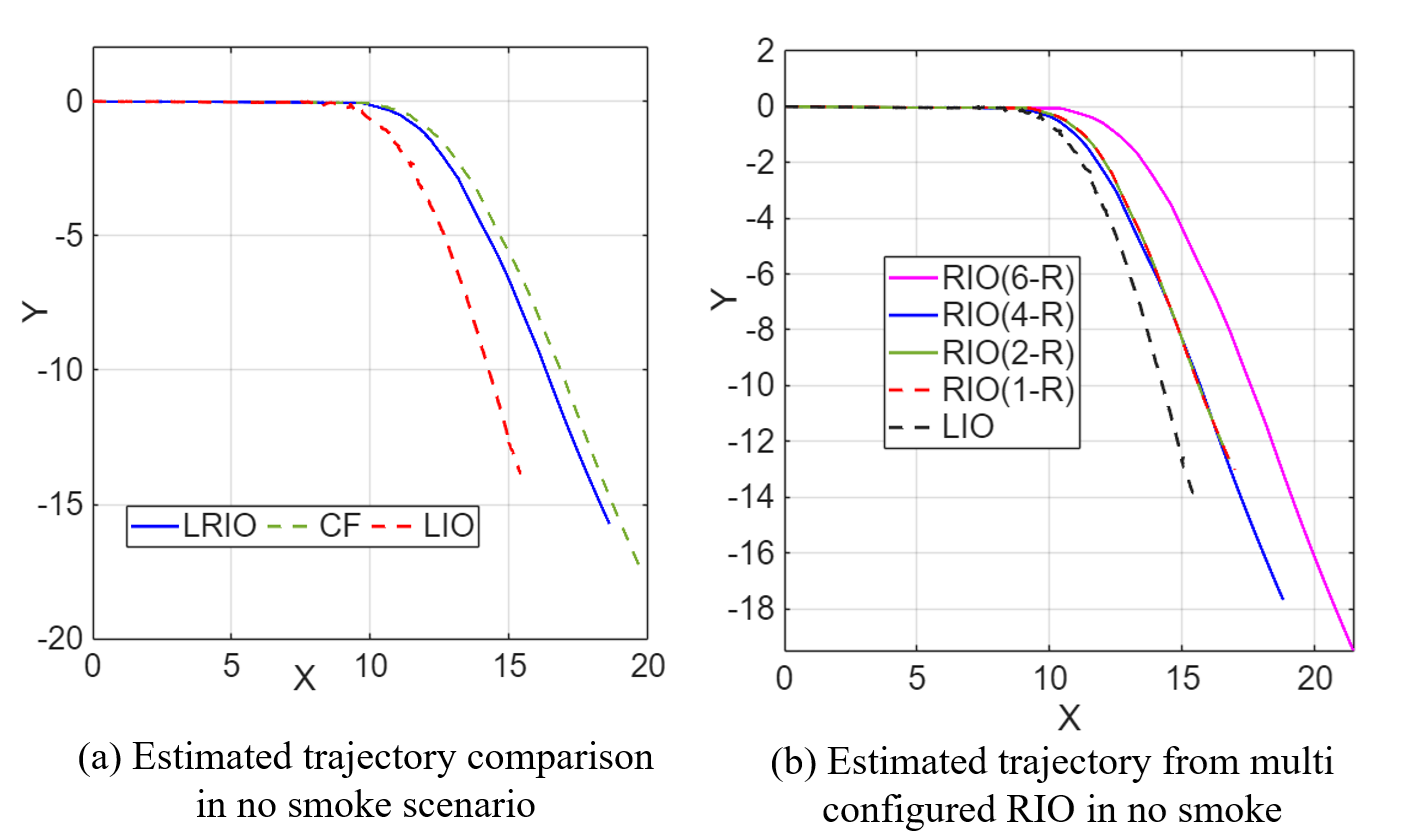}
    \caption{In this context, LIO is treated as the ground-truth reference, while acknowledging that each sensing modality has its own limitations. Notably, the Lidar radar inertial odometry~(LRIO) closely follows the path recorded during the pre-experimental measurements.}
    \label{fig:DLRIO_based_trajectory_no_smoke}
\end{figure*}

In Environment-1~( Fig.~\ref{fig:Environment_1}), the robot navigated an 'L'-shaped trajectory. In the absence of smoke, all sensors operate optimally. As indicated, the proposed LRIO, presented in Fig.~\ref{fig: Concept_figure_LRIO}, uses four distinct radar configuration setups as individual sensor modules. Each configuration's unique heading and field of view produce distinct pointclouds, resulting in separate RIO estimates presented in Fig.~\ref{fig:DLRIO_based_trajectory_no_smoke}(b). Consequently, these estimates vary across configurations functioning as independent modules in this sensor fusion algorithm. Our proposed algorithm~(Fig.~\ref{fig: Concept_figure_LRIO}) uses an initial fusion layer that evaluates a set of $n=5$ likelihood functions, where each function is a module that consists of an extended Kalman Filter, in which $ n-1=4$ sensor configurations are fused in each node. For example, filter-1 fuses measurements from RIO-2, RIO-4, RIO-6 and LIO, while filter-2 integrates RIO-1, RIO-4, RIO-6 and LIO. The first fusion layer produces five distinct pose estimates as illustrated in Fig.~\ref{fig: Nosmk_filter5_all_odom}~(b) by combining individual sensor information within their respective likelihood functions.
All filters are used to generate residuals and were initialized with identical state vectors since the same robot kinematics are used in all filters inside the proposed LRIO framework, with the initial state set to $\mathcal{X}_0 = \mathbf{0}_{4\times1}$ for the filter bank, and the initial state covariance of each likelihood filter is chosen as $\Sigma_0 = \boldsymbol{I}_{4\times4}$. The process noise covariance matrix Q, defined as $(Q=ww^\top)$, while $w$ denotes the process noise and $\nu$ denotes the measurement noise and noise covariance R, defined as $R=\nu\nu^\top$ considered identical for all filters, with $Q= 100\,\boldsymbol{I}_{4\times4}$ and $R= \boldsymbol{I}_{20\times20}$. These estimates from the first layer filter bank are used to generate residuals, which are compared using adaptive thresholding, or by analyzing the regions where residuals associated with the MLEs exist, as shown in Figs.\ref{fig: Nosmk_filter_1_2}-\ref{fig: Nosmk_filter5_all_odom}(a). Each likelihood function can generate a highest likelihood estimate for a few time samples using autonomous adaptive thresholding. The adaptive thresholding process analyzes the residual vectors, which are derived from a reference filter, theoretically explained in Section~\ref{sec: Residue_generation}. 
These residual vectors are the basis for a statistical analysis, assuming all the residuals per sample follow a normal distribution with non-zero mean. The lower and upper bounds of this proposed adaptive residual thresholding have a tuning parameter $\mathcal{K}$, which controls how wide or narrow the thresholding region per sample. Importantly, the range does not have to stay fixed,it can vary for each sample. By adjusting this range dynamically, the method tries to identify the region where the likelihood estimates are most likely to correspond to the Maximum Likelihood Estimate (MLE). 

This proposed thresholding algorithm~(sec.~\ref{sec: adaptive threshold}) aims to determine the highest likelihood for each instance.  The residue corresponding to this filter lies within this bound, yielding the logical outcome 1 or 0 for each first-layer filter estimate. According to this logical expression, the final integration layer integrates them optimally. Fig.~\ref{fig:LRIO_based_lcal_radar_map_compared_LIO}  illustrates the resulting pose estimates obtained through the final optimal integration under smoke-free conditions. These figures indicate the periods during which specific residuals consistently lie within the threshold bounds, signifying that the corresponding filter yields trustworthy state estimations throughout those intervals. For example, in Fig.~\ref{fig: Nosmk_filter5_all_odom}(a), the residual $R_{e_5}$, is shown by the green line, and stays within the threshold. This suggests that the likelihood function (LF-5) provides state estimates that are more consistent with the MLEs compared to the pose estimates of other filters, either for that specific time instance or over the entire duration. Consequently, the region $\Delta_k$ determines whether the parameter $\delta_{l_k}$ takes the value $0$ or $1$, where $l$ represents the total number of filters in the first layer. For example, if $l=5$, then $\delta_{5_k}$ corresponds to the fifth estimator in the filter bank, and this determines which estimates are forwarded to the final fusion layer. Ultimately, fusing these estimates yields the final optimal state estimates, which closely agree with the ground truth, as illustrated in Figs.~\ref{fig:DLRIO_based_trajectory_no_smoke}(a) and~\ref{fig:LRIO_based_lcal_radar_map_compared_LIO}. In addition, the performance of the centralized fusion approach in Eq.~\ref{eq:estimators_n_sensors} is compared with both LIO and the proposed method in Fig.~\ref{fig:DLRIO_based_trajectory_no_smoke}(a). The overall computational complexity of the proposed LRIO framework is evaluated in terms of execution time per update. The first-layer filter bank computation, residual generation, statistical selection of the maximum likelihood estimate (MLE), and final integration of the selected MLEs within the isolation layer require a total of 0.008225 s per update. Since each of the four RIO formulations consists of two processing stages per configuration, the complete RIO framework requires 0.0658 s per update. This computational cost demonstrates that the proposed decentralized fusion framework for LiDAR and radars is suitable for real-time operation.
\begin{table}[H]
\centering
\caption{Quantitative localization error statistics for the evaluated methods.}
\label{tab:Error_Statistics}
\begin{tabular}{lccc}
\hline
Method & Mean Error & Std. Error & RMSE \\
\hline
Proposed (RETOIF) & 1.6393 & 1.4270 & 2.1733 \\
CF (LRIO)         & 2.1851 & 2.0453 & 2.9929 \\
1-RIO             & 0.8599 & 0.4497 & 0.9704 \\
2-RIO             & 0.9004 & 0.4996 & 1.0297 \\
4-RIO             & 1.4840 & 1.7988 & 2.3319 \\
6-RIO             & 2.9086 & 2.7226 & 3.9839 \\
\hline
\end{tabular}
\end{table}
\subsection{Quantitative Localization Error Analysis}
The quantitative results in Table~\ref{tab:Error_Statistics} are consistent with the observed localization error trends. Among the individual RIO configurations, 1-RIO and 2-RIO achieve the lowest RMSE values of 0.9704 and 1.0297, respectively, while the error increases to 2.3319 for 4-RIO and 3.9839 for 6-RIO. This confirms that increasing the number of radar inputs does not necessarily improve localization accuracy when additional measurements contain inconsistent or degraded information. The conventional centralized fusion method, CF (LRIO), yields an RMSE of 2.9929, whereas the proposed RETOIF reduces the RMSE to 2.1733, corresponding to an improvement of approximately 27.4\%. RETOIF also reduces the mean error from 2.1851 to 1.6393 and the standard error from 2.0453 to 1.4270 compared with CF (LRIO). Although 1-RIO and 2-RIO achieve lower absolute errors in this smoke-free experiment, RETOIF provides a more robust heterogeneous fusion solution by selectively limiting the influence of inconsistent sensor measurements through adaptive likelihood evaluation, residual based consistency assessment, and fault isolation.

\subsection{Experimental Evaluation in Environment-2}
\begin{figure*}[h]
    \centering
    \begin{subfigure}{0.32\textwidth}
        \centering
        \includegraphics[width=\textwidth]{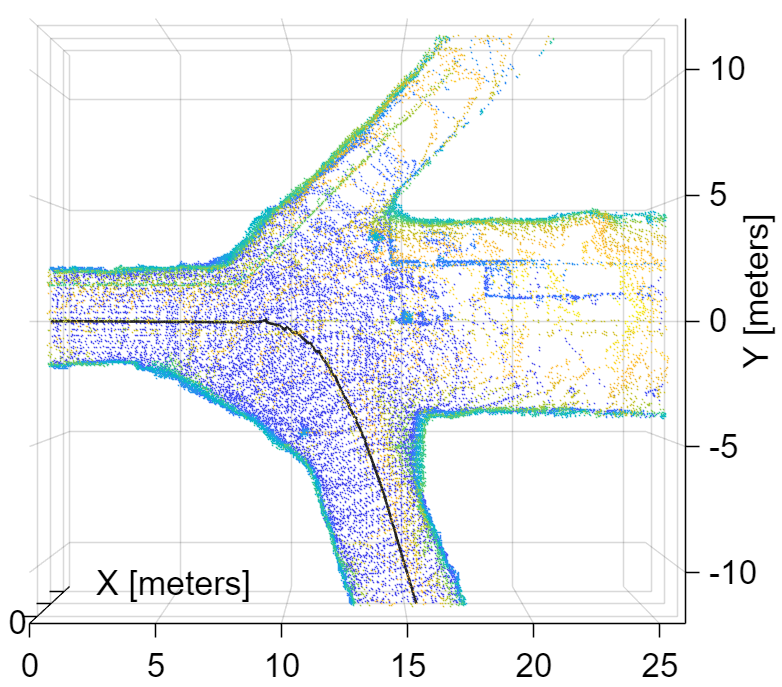}
        \caption{No smoke}
        \label{fig:GT_LIO}
    \end{subfigure}
    \hfill
    \begin{subfigure}{0.32\textwidth}
        \centering
        \includegraphics[width=\textwidth]{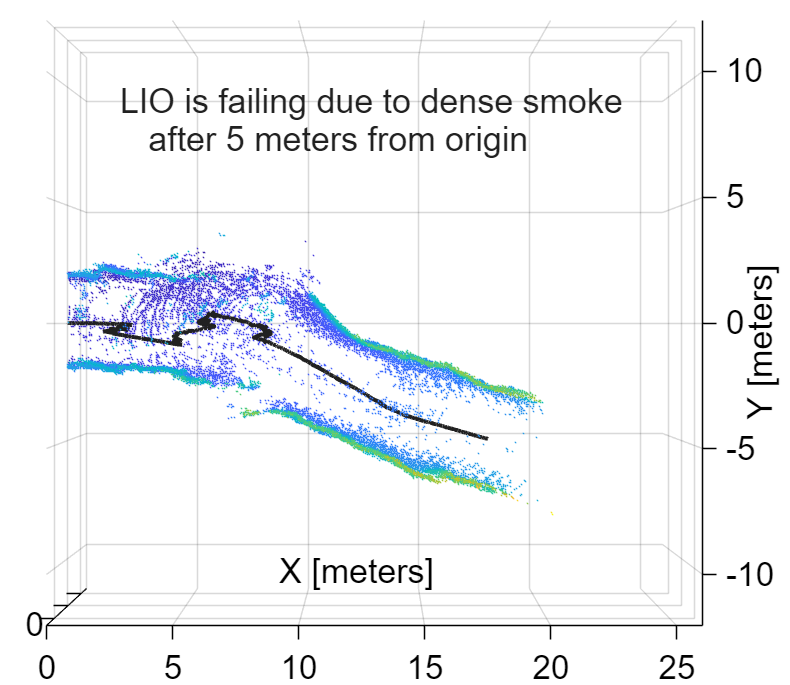}
        \caption{Dense smoke (Case-1).}
        \label{fig:case1_Smk_LIO}
    \end{subfigure}
    \hfill
    \begin{subfigure}{0.32\textwidth}
        \centering
        \includegraphics[width=\textwidth]{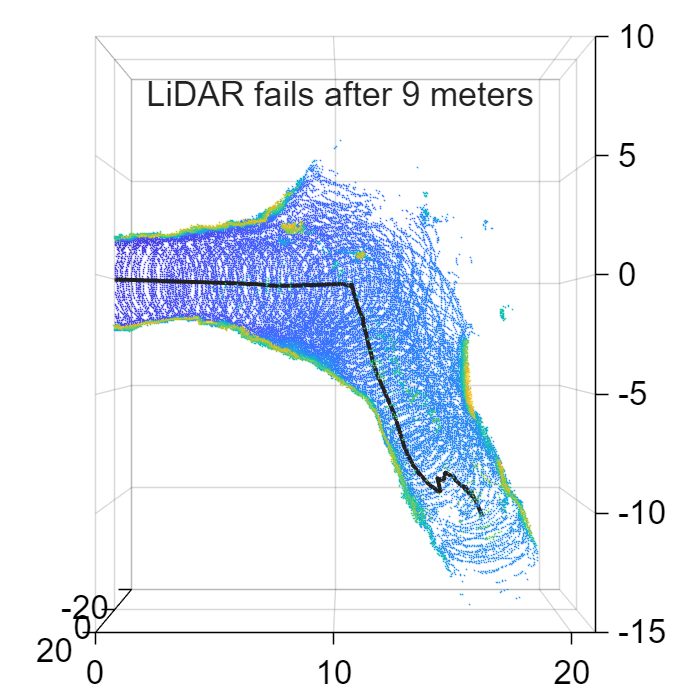}
        \caption{Light smoke (Case-2).}
        \label{fig:case2_Smk_LIO}
    \end{subfigure}
    \caption{Comparison of LiDAR-based localization and mapping performance in the experimental SubT tunnel under different visibility conditions: (a) no smoke, (b) dense smoke (Case-1), and (c) light smoke (Case-2). The results illustrate the progressive degradation of LiDAR-based localization and mapping in the presence of smoke.}
    \label{fig:three_cases}
\end{figure*}

\begin{figure*}[h]
\centering
\includegraphics[width=1.0\textwidth]{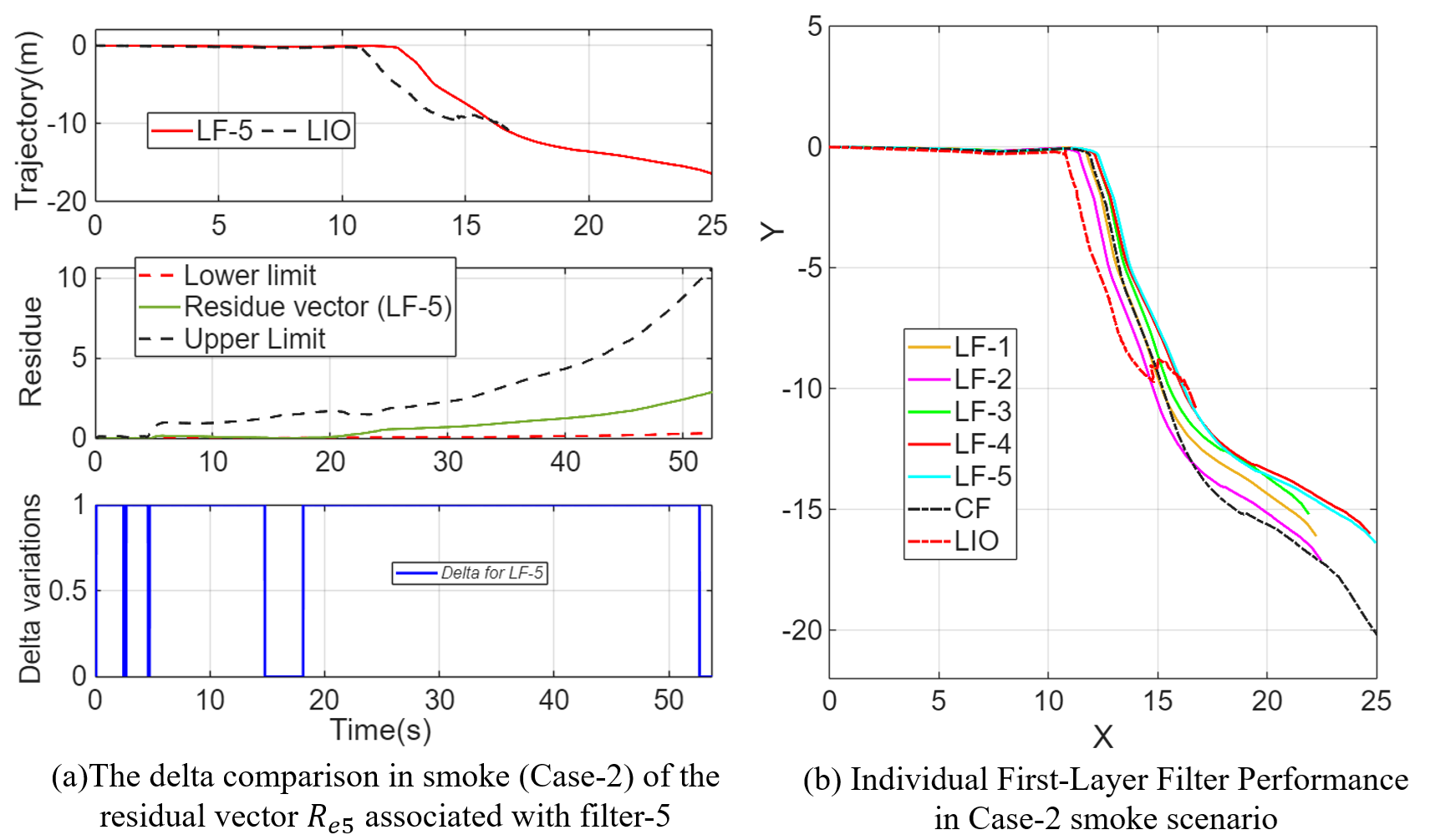}
\caption{Residual and delta evaluation of LF-5, together with the trajectory performance of the individual first-layer filter estimates under smoke conditions (Case-1).}
\label{fig: case_2_smk_dlta_filt_5_odom}
\end{figure*}

\begin{figure*}[h]
\centering
\includegraphics[width=1.0\textwidth]{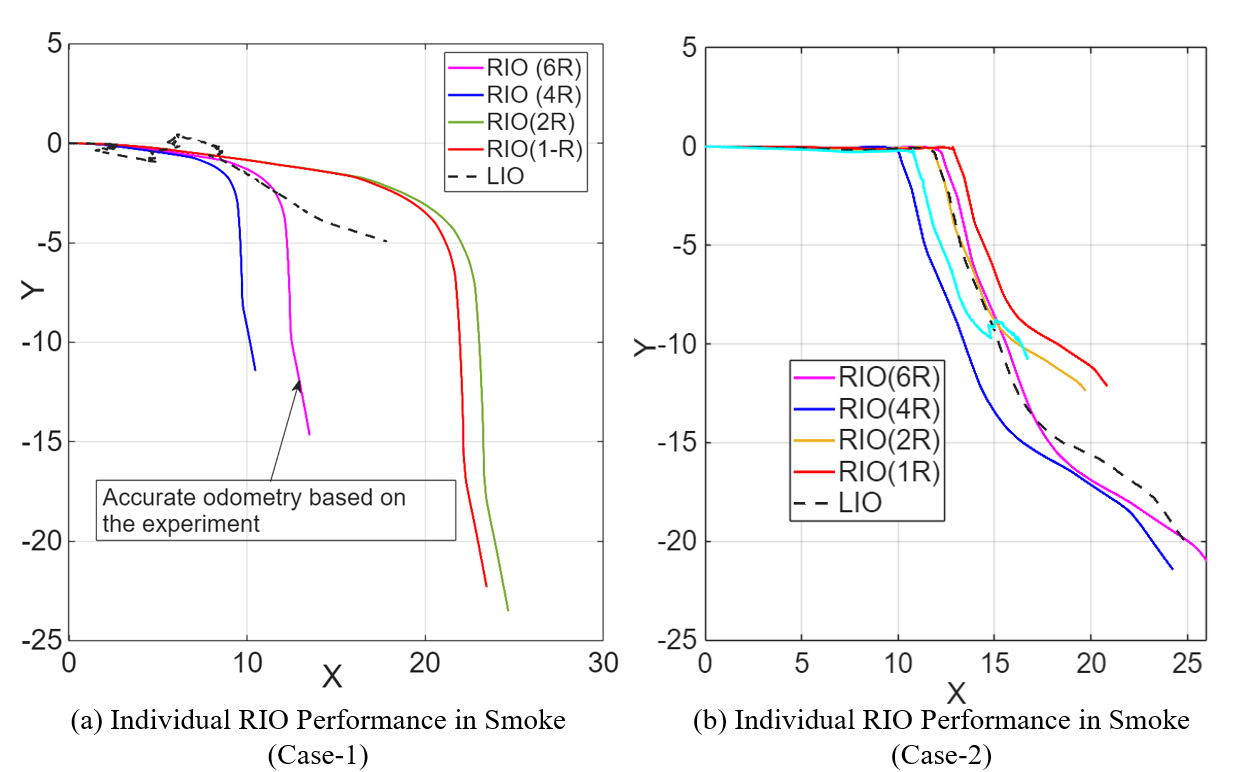}
\caption{Comparison of all RIO configurations with LIO in a smoke-filled tunnel for Case-1 and Case-2 scenarios}
\label{All_Conf_RIO_smk_1_2}
\end{figure*}

Environment-2~(Fig.~\ref{fig:Environment_2}), the robot followed the same trajectory, covering nearly identical distances as in Environment-1 (\ref{sec: Env1_results}). The experiments were conducted in an underground tunnel where smoke conditions varied in density, transitioning from denser accumulation to light dispersion over the tunnels over time. During experimental evaluation, to replicate visually degraded subterranean environments, smoke was introduced using a theatrical fog machine together with a standard water-based fog fluid. The fog machine was positioned inside the tunnel and operated at a constant output level during each trial. To systematically investigate the impact of smoke on localization performance, two experiments were conducted in Environment-2 under different smoke density conditions: dense and light smoke, denoted as Case-1 and Case-2, respectively. In both cases, the datasets were collected at the same location while maintaining identical geometric conditions. The performance of radar-inertial odometry across various configurations under dense and light smoke conditions is presented in Figs.~\ref{All_Conf_RIO_smk_1_2}(a) and (b), corresponding to Case-1 and Case-2, respectively. Furthermore, the performance of LiDAR-based odometry under different smoke dispersion conditions is illustrated in Figs.~\ref{fig:GT_LIO}, Fig.~\ref{fig:case1_Smk_LIO} and Fig.~\ref{fig:case2_Smk_LIO}.
\begin{figure*}[h]
\centering
\includegraphics[width=1.0\textwidth]{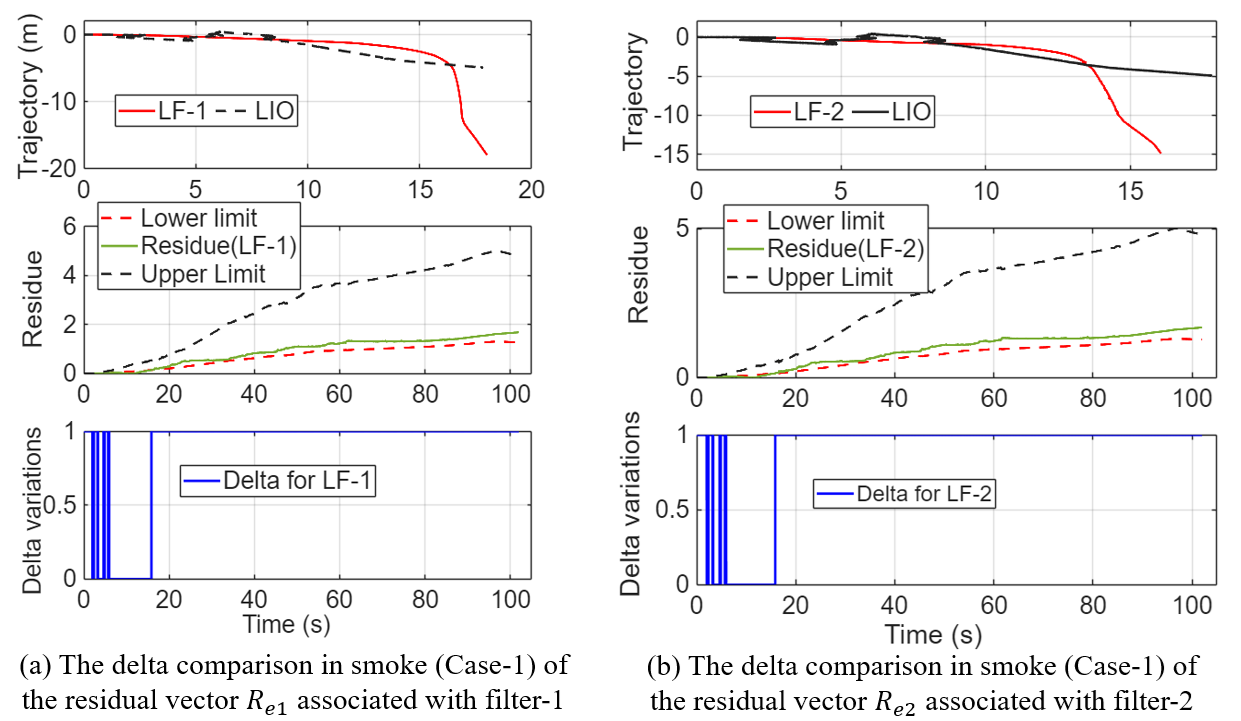}
\caption{Performance analysis of hypothesis testing based on the adaptive selection of the threshold region for the first-layer filters in the Case-2 smoke test scenario within Environment-2}
\label{fig: case_1_smk_dlta_filt_1_2}
\end{figure*}
\begin{figure*}[h]
\centering
\includegraphics[width=1.0\textwidth]{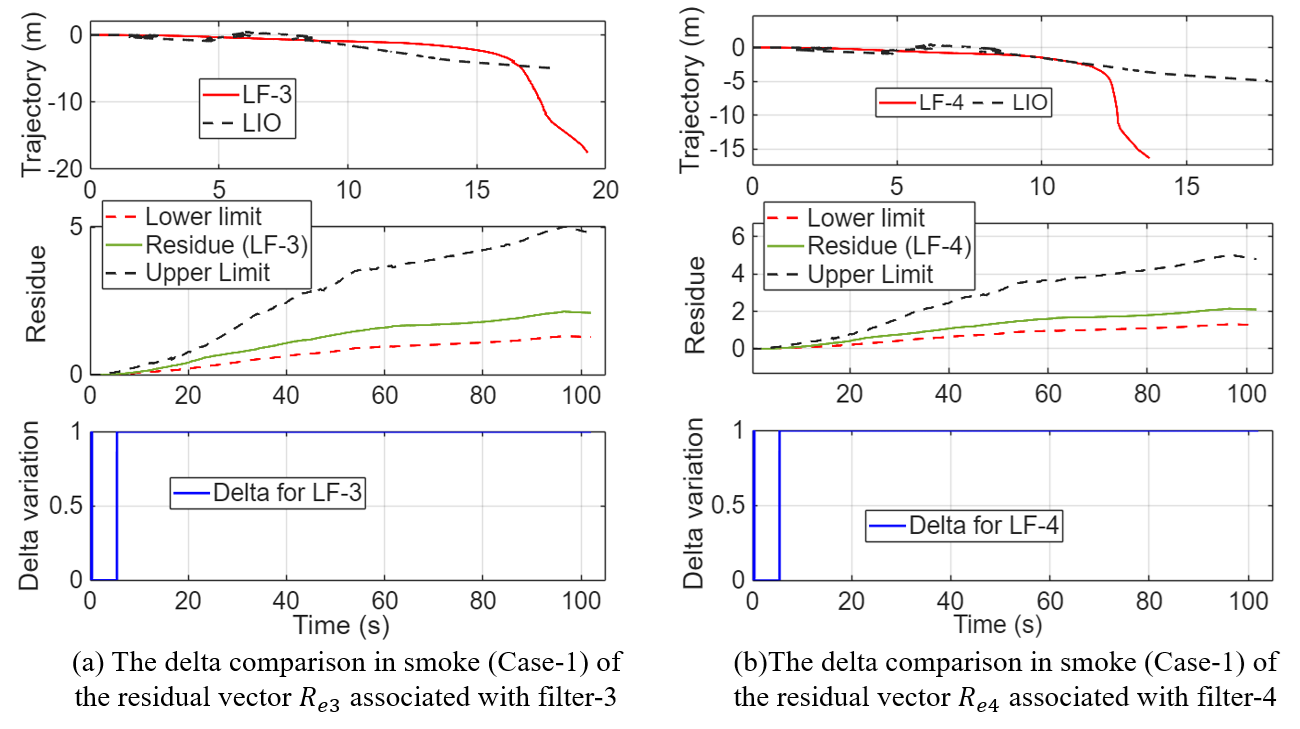}
\caption{Performance analysis of hypothesis testing based on the adaptive selection of the threshold region for the first-layer filters in the Case-2 smoke test scenario within Environment-2}
\label{fig:case_1_smk_dlta_filt_3_4}
\end{figure*}
\begin{figure*}[h]
\centering
\includegraphics[width=1.0\textwidth]{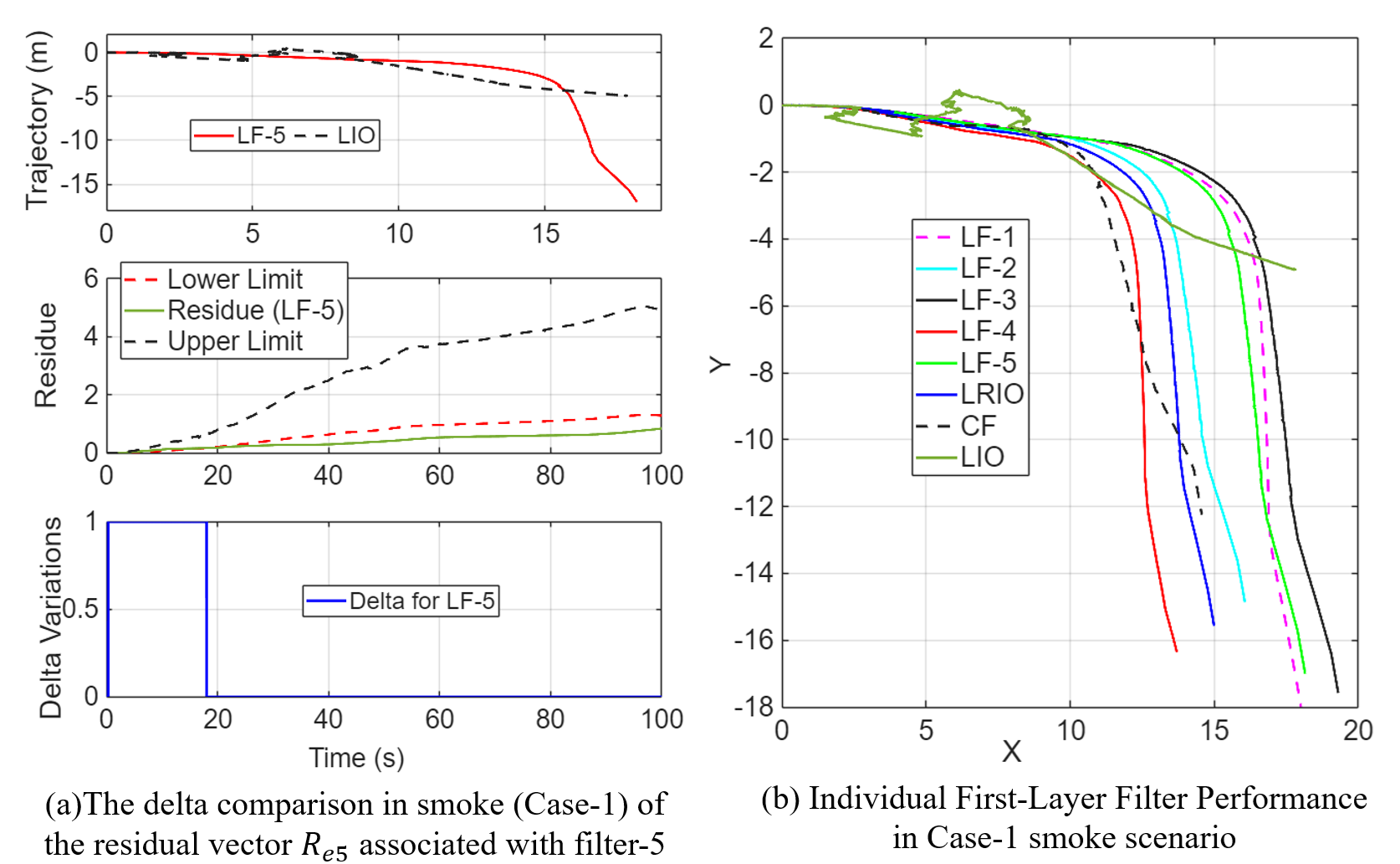}
\caption{Performance analysis of hypothesis testing based on the adaptive selection of the threshold region for the first-layer filters in the Case-2 smoke test scenario within Environment-2}
\label{fig: case_1_smk_dlta_filt_5_odom}
\end{figure*}
These first-layer estimated states~(detailed in sec.~\ref{sec: Residue_generation}) generate a set of residue vectors at time step $k$. If the residual error patterns remain within the predefined threshold bounds, the corresponding filters are regarded as the most likely estimates for that time step. Over time, these MLEs are optimally integrated into the final layer as described in Sec.~\ref{sec: Hypothesis test}, ensuring the selection of the most reliable state estimates at each interval. The pose estimates obtained from the first-layer filters, corresponding to the likelihood functions for Case-1 and Case-2, are illustrated in Fig.~\ref{fig: case_1_smk_dlta_filt_5_odom}(b) and \ref{fig: case_2_smk_dlta_filt_5_odom}(b).

\begin{figure*}[h]
\centering
\includegraphics[width=1.0\textwidth]{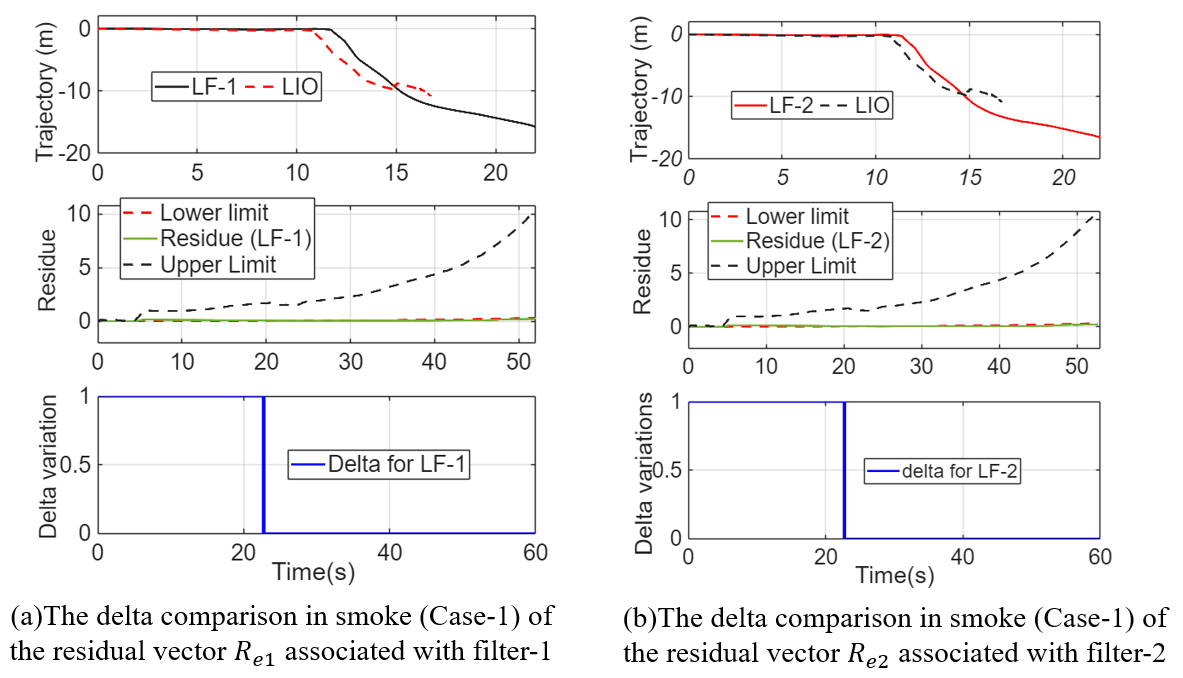}
\caption{Residual-based hypothesis evaluation for the two local filters under smoke conditions (Case-1), showing the estimated trajectories, residual bounds, and corresponding delta variations for LF-1 and LF-2.}
\label{fig: case_2_smk_dlta_filt_1_2}
\end{figure*}

\begin{figure*}[h]
\centering
\includegraphics[width=1.0\textwidth]{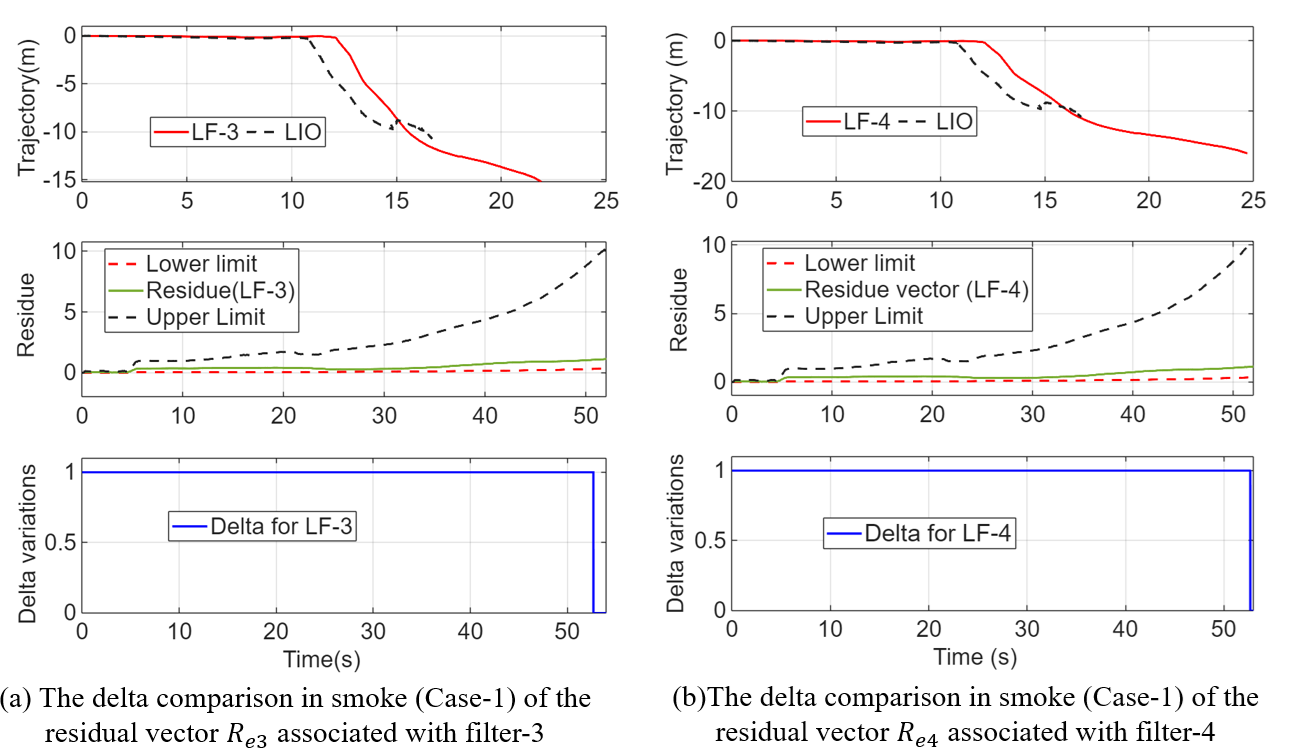}
\caption{Residual based hypothesis evaluation for LF-3 and LF-4 under smoke conditions (Case-1), showing trajectory estimates, residual bounds, and corresponding delta variations.}
\label{fig: case_2_smk_dlta_filt_3_4}
\end{figure*}

Similar to Environment-1, Figs.~\ref{fig: case_1_smk_dlta_filt_1_2}-\ref{fig: case_1_smk_dlta_filt_5_odom}(a) present the $\Delta$-based comparison of the residual vector $R_{e_3}$ associated with the likelihood function $f_3$, which fuses RIO configurations comprising one, two, four, and six radars. Under highly dense smoke conditions, this likelihood function does not incorporate the LIO dataset. Consequently, the corresponding residual vector remains within the predefined threshold bounds, and the estimates obtained from LF-3 can therefore be regarded as the maximum likelihood estimates (MLEs) for this experiment. Similar behaviour is observed in Figs.~\ref{fig: case_1_smk_dlta_filt_1_2}-\ref{fig: case_2_smk_dlta_filt_5_odom} which depict the $\Delta_k$ for each likelihood function within the adaptive threshold limits. These MLEs are subsequently integrated to produce the final estimates of the proposed framework.

\begin{figure*}[h]
    \centering
    \begin{subfigure}{0.48\textwidth}
        \centering
        \includegraphics[width=\textwidth]{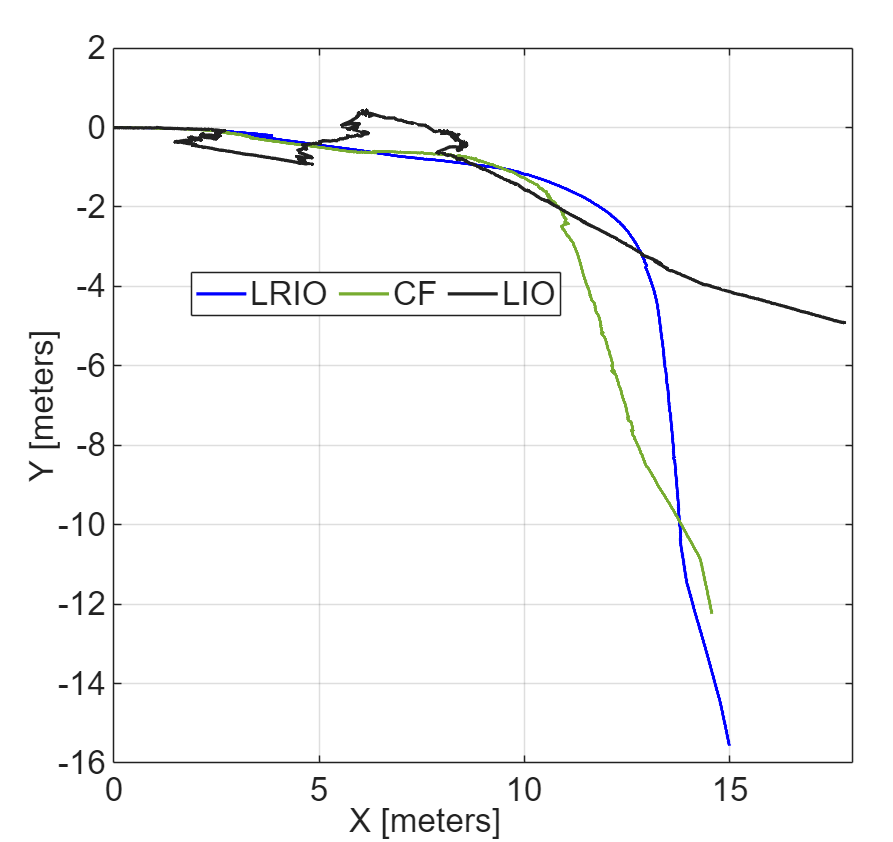}
        \caption{LRIO performance under dense smoke conditions in the Case-1 scenario}
        \label{fig:LRIO_CASE_1}
    \end{subfigure}
    \hfill
    \begin{subfigure}{0.48\textwidth}
        \centering
        \includegraphics[width=\textwidth]{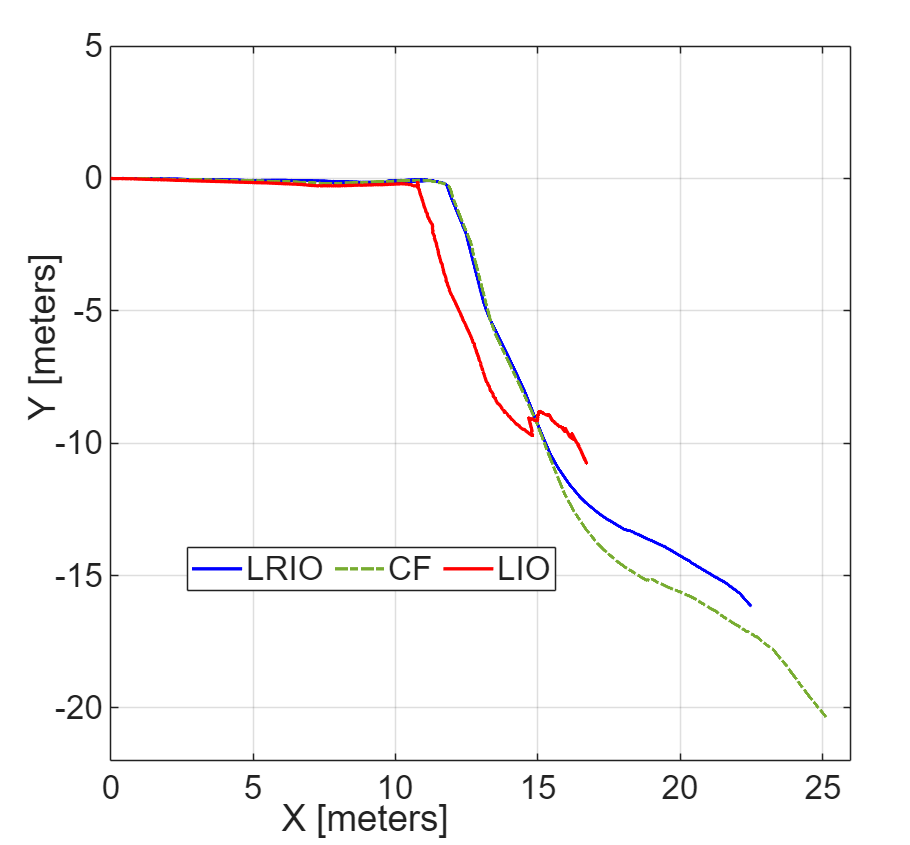}
        \caption{LRIO performance under dense smoke conditions in the Case-2 scenario}
        \label{fig:LRIO_CASE_2}
    \end{subfigure}
    \caption{Comparison of trajectory estimation performance under smoke-degraded conditions. (a) Performance of the proposed LRIO framework compared with LIO and centralized fusion in Case-1. (b) Performance of the proposed LRIO framework under dense-smoke (Case-1) and light-smoke (Case-2) conditions.}
    \label{fig:Case1_2_LRIO_Traj_Est}
\end{figure*}
Figs.~\ref{fig:LRIO_CASE_1} and ~\ref{fig:LRIO_CASE_2} illustrate the extent to which smoke disrupts LIO-based perception, demonstrating both the challenges posed by such environments and the necessity of robust sensor fusion techniques to maintain accurate odometry in adverse conditions. As illustrated in Fig.~\ref{fig: Concept_figure_LRIO}, the proposed RETOIF hierarchical fusion strategy generated multiple likelihood estimates during the first-layer fusion for the Case-1 and Case-2 smoke-test experiments conducted in the Environment-2 scenario. Since this decentralized fusion integrates multiple radar sensors, the resulting trajectory estimates are predominantly influenced by radar measurements. This influence becomes particularly significant in the absence of reliable LIO data, affecting the precision and accuracy of the robot’s localization, especially with respect to trajectory magnitudes. Figs.~\ref{fig:LRIO_CASE_1} and ~\ref{fig:LRIO_CASE_2} demonstrate that the proposed architecture achieves accurate trajectory estimation when compared against ground truth measurements obtained using a measuring tape for both straight and curved segments of trajectories.

Extensive experimental validation in subterranean tunnel environments confirms the ability of the proposed approach to address the limitations of centralized frameworks under both nominal and degraded sensing conditions.
\section{Conclusion}
This study represents a resilient decentralized LiDAR Radar Inertial Odometry (LRIO) framework based on the Robust Error-Tolerant Optimal Information Filter (RETOIF) for autonomous localization in GPS-denied and visually degraded subterranean environments. By integrating complementary LiDAR, radar, and inertial sensing modalities through adaptive residual thresholding, likelihood-based consistency evaluation, and selective sensor fusion, the proposed framework enhances robustness against inconsistent measurements and temporary sensor degradation. Experimental results in the smoke-free environment demonstrate that the proposed RETOIF framework achieves a localization RMSE of 2.1733, compared with 2.9929 for conventional centralized fusion (CF-LRIO), corresponding to a 27.4\% reduction in localization error. Additionally, the proposed method reduces the mean localization error from 2.1851 to 1.6393 and the reported standard error from 2.0453 to 1.4270. Although the individual 1-RIO and 2-RIO configurations achieve lower RMSE values of 0.9704 and 1.0297, respectively, over the short 26~m experimental trajectory, RETOIF provides a heterogeneous fusion strategy that adaptively selects reliable pose estimates rather than relying on a single sensor configuration. Further experimental evaluations under light and dense smoke conditions demonstrate the framework's ability to maintain trajectory estimation despite substantial degradation of LiDAR inertial odometry. The adaptive residual thresholding mechanism facilitates the identification and exclusion of inconsistent estimates, while the integration of multiple radar configurations provides complementary environmental coverage under adverse sensing conditions. These findings highlight the potential of decentralized fusion to improve localization resilience when individual sensing modalities become unreliable.
\section{Declarations}
\subsection{Conflicts of Interest}
The authors have no conflicts of interest or competing interests with any related parties.

\subsection{Funding}
This work has been partially funded by the European Union’s Horizon Europe Research and Innovation Programme under Grant Agreement No.101138451 PERSEPHONE and partially by the research project SP14, ‘Autonomous Drones for Underground Mining Operations’, within the academic programme of the Sustainable Underground Mining (SUM) project, which LKAB and the Swedish Energy Agency jointly finance.

\subsection{CRediT authorship contribution statement}
\textbf{Moumita Mukherjee:} Conceptualisation, Methodology, Software, Validation, Investigation, Data curation, Formal analysis, Writing original draft, Writing review editing and Visualisation.
\textbf{Avijit Banerjee:} Formal analysis, Investigation, Visualisation, Writing - Review \& Editing, Supervision. 
\textbf{Anton Koval:} Formal analysis, Data Curation, Project administration, Writing - Review \& Editing, Supervision.
\textbf{George Nikolakopoulos:} Conceptualisation, Formal analysis, Project administration, Funding acquisition, Resources, Writing - Review \& Editing, Supervision.

\subsection{Data Availability}
Data will be made available on reasonable request.

\appendix

\bibliographystyle{elsarticle-num}
\bibliography{MLRIO_ref.bib}

\begin{thebibliography}{10}
\expandafter\ifx\csname url\endcsname\relax
  \def\url#1{\texttt{#1}}\fi
\expandafter\ifx\csname urlprefix\endcsname\relax\def\urlprefix{URL }\fi
\expandafter\ifx\csname href\endcsname\relax
  \def\href#1#2{#2} \def\path#1{#1}\fi

\bibitem{lindqvist2022compra}
B.~Lindqvist, C.~Kanellakis, S.~S. Mansouri, A.-a. Agha-mohammadi, G.~Nikolakopoulos, Compra: A compact reactive autonomy framework for subterranean mav based search-and-rescue operations, Journal of Intelligent \& Robotic Systems 105~(3) (2022) 49.
\newblock \href {https://doi.org/https://doi.org/10.1007/s10846-022-01665-6} {\path{doi:https://doi.org/10.1007/s10846-022-01665-6}}.

\bibitem{agha2021nebula}
A.~Agha, K.~Otsu, B.~Morrell, D.~D. Fan, R.~Thakker, A.~Santamaria-Navarro, S.-K. Kim, A.~Bouman, X.~Lei, J.~Edlund, et~al., Nebula: Quest for robotic autonomy in challenging environments; team costar at the darpa subterranean challenge, arXiv preprint arXiv:2103.11470 (2021).
\newblock \href {https://doi.org/https://doi.org/10.48550/arXiv.2103.11470} {\path{doi:https://doi.org/10.48550/arXiv.2103.11470}}.

\bibitem{taleb2025automotive}
M.~A. Taleb, G.~Korsoveczki, G.~Husi, Automotive navigation for mobile robots: Comprehensive review, Results in Engineering (2025) 105837\href {https://doi.org/https://doi.org/10.1016/j.rineng.2025.105837} {\path{doi:https://doi.org/10.1016/j.rineng.2025.105837}}.

\bibitem{kim2023empirical}
J.~Kim, B.-j. Park, J.~Kim, Empirical analysis of autonomous vehicle’s lidar detection performance degradation for actual road driving in rain and fog, Sensors 23~(6) (2023) 2972.
\newblock \href {https://doi.org/DOI: 10.3390/s23062972} {\path{doi:DOI: 10.3390/s23062972}}.

\bibitem{piroli2023energy}
A.~Piroli, V.~Dallabetta, J.~Kopp, M.~Walessa, D.~Meissner, K.~Dietmayer, Energy-based detection of adverse weather effects in lidar data, IEEE Robotics and Automation Letters 8~(7) (2023) 4322--4329.
\newblock \href {https://doi.org/10.1109/LRA.2023.3282382} {\path{doi:10.1109/LRA.2023.3282382}}.

\bibitem{park2025lidar}
J.-i. Park, S.~Jo, H.-T. Seo, J.~Park, Lidar denoising methods in adverse environments: A review, IEEE Sensors Journal (2025).
\newblock \href {https://doi.org/10.1109/JSEN.2025.3526175} {\path{doi:10.1109/JSEN.2025.3526175}}.

\bibitem{vargas2020weather}
J.~R. Vargas~Rivero, T.~Gerbich, V.~Teiluf, B.~Buschardt, J.~Chen, Weather classification using an automotive lidar sensor based on detections on asphalt and atmosphere, Sensors 20~(15) (2020) 4306.
\newblock \href {https://doi.org/10.3390/s20154306} {\path{doi:10.3390/s20154306}}.

\bibitem{chen2023drio}
H.~Chen, Y.~Liu, Y.~Cheng, Drio: Robust radar-inertial odometry in dynamic environments, IEEE Robotics and Automation Letters (2023).
\newblock \href {https://doi.org/10.3390/drones8050197} {\path{doi:10.3390/drones8050197}}.

\bibitem{doer2020ekf}
C.~Doer, G.~F. Trommer, An ekf based approach to radar inertial odometry, in: 2020 IEEE International Conference on Multisensor Fusion and Integration for Intelligent Systems (MFI), IEEE, 2020, pp. 152--159.
\newblock \href {https://doi.org/10.1109/MFI49285.2020.9235254} {\path{doi:10.1109/MFI49285.2020.9235254}}.

\bibitem{zhuang20234d}
Y.~Zhuang, B.~Wang, J.~Huai, M.~Li, 4d iriom: 4d imaging radar inertial odometry and mapping, IEEE Robotics and Automation Letters 8~(6) (2023) 3246--3253.
\newblock \href {https://doi.org/10.1109/LRA.2023.3266669} {\path{doi:10.1109/LRA.2023.3266669}}.

\bibitem{doer2021yaw}
C.~Doer, G.~F. Trommer, Yaw aided radar inertial odometry using manhattan world assumptions, in: 2021 28th Saint Petersburg International Conference on Integrated Navigation Systems (ICINS), IEEE, 2021, pp. 1--9.
\newblock \href {https://doi.org/10.23919/ICINS43216.2021.9470842} {\path{doi:10.23919/ICINS43216.2021.9470842}}.

\bibitem{herraez2024radar}
D.~C. Herraez, M.~Zeller, L.~Chang, I.~Vizzo, M.~Heidingsfeld, C.~Stachniss, Radar-only odometry and mapping for autonomous vehicles, in: 2024 IEEE International Conference on Robotics and Automation (ICRA), IEEE, 2024, pp. 10275--10282.
\newblock \href {https://doi.org/10.3390/s23135779} {\path{doi:10.3390/s23135779}}.

\bibitem{nissov2024degradation}
M.~Nissov, N.~Khedekar, K.~Alexis, Degradation resilient lidar-radar-inertial odometry, arXiv preprint arXiv:2403.05332 (2024).
\newblock \href {https://doi.org/10.1109/ICRA57147.2024.10611444} {\path{doi:10.1109/ICRA57147.2024.10611444}}.

\bibitem{barlow1949doppler}
E.~J. Barlow, Doppler radar, Proceedings of the IRE 37~(4) (1949) 340--355.
\newblock \href {https://doi.org/10.1109/JRPROC.1949.231638} {\path{doi:10.1109/JRPROC.1949.231638}}.

\bibitem{choi2023msc}
M.~Choi, S.~Yang, S.~Han, Y.~Lee, M.~Lee, K.~H. Choi, K.-S. Kim, Msc-rad4r: Ros-based automotive dataset with 4d radar, IEEE Robotics and Automation Letters 8~(11) (2023) 7194--7201.
\newblock \href {https://doi.org/10.1109/LRA.2023.3307005} {\path{doi:10.1109/LRA.2023.3307005}}.

\bibitem{zhang20234dradarslam}
J.~Zhang, H.~Zhuge, Z.~Wu, G.~Peng, M.~Wen, Y.~Liu, D.~Wang, 4dradarslam: A 4d imaging radar slam system for large-scale environments based on pose graph optimization, in: 2023 IEEE International Conference on Robotics and Automation (ICRA), IEEE, 2023, pp. 8333--8340.
\newblock \href {https://doi.org/10.1109/LRA.2023.3307005} {\path{doi:10.1109/LRA.2023.3307005}}.

\bibitem{mukherjee2026resilient}
M.~Mukherjee, M.~Nor{\'e}n, A.~Koval, A.~Banerjee, G.~Nikolakopoulos, Resilient multi-radar inertial odometry with adaptive bias correction for localization in smoke-filled subterranean environments, Journal of Intelligent \& Robotic Systems 112~(2) (2026) 31.

\bibitem{teng2024multi}
X.~Teng, Z.~Shen, L.~Huang, H.~Li, W.~Li, Multi-sensor fusion based wheeled robot research on indoor positioning method, Results in Engineering 22 (2024) 102268.
\newblock \href {https://doi.org/https://doi.org/10.1016/j.rineng.2024.102268} {\path{doi:https://doi.org/10.1016/j.rineng.2024.102268}}.

\bibitem{li2024continuous}
S.~Li, X.~Li, S.~Chen, Y.~Zhou, S.~Wang, Continuous-time lidar/imu/radar odometry for accurate and smooth state estimation via tightly coupled integration, IEEE Internet of Things Journal (2024).
\newblock \href {https://doi.org/10.1109/JIOT.2024.3450654} {\path{doi:10.1109/JIOT.2024.3450654}}.

\bibitem{zhao2021super}
S.~Zhao, H.~Zhang, P.~Wang, L.~Nogueira, S.~Scherer, Super odometry: Imu-centric lidar-visual-inertial estimator for challenging environments, in: 2021 IEEE/RSJ International Conference on Intelligent Robots and Systems (IROS), IEEE, 2021, pp. 8729--8736.
\newblock \href {https://doi.org/10.1109/IROS51168.2021.9635862} {\path{doi:10.1109/IROS51168.2021.9635862}}.

\bibitem{kim2020extended}
T.~Kim, T.-H. Park, Extended kalman filter (ekf) design for vehicle position tracking using reliability function of radar and lidar, Sensors 20~(15) (2020) 4126.
\newblock \href {https://doi.org/https://doi.org/10.3390/s20154126} {\path{doi:https://doi.org/10.3390/s20154126}}.

\bibitem{mukherjee2021decentralized}
M.~Mukherjee, A.~Banerjee, A.~Papadimitriou, S.~S. Mansouri, G.~Nikolakopoulos, A decentralized sensor fusion scheme for multi sensorial fault resilient pose estimation, Sensors 21~(24) (2021) 8259.
\newblock \href {https://doi.org/10.3390/s21248259} {\path{doi:10.3390/s21248259}}.

\bibitem{sun2003multi}
S.~Sun, Multi-sensor optimal information fusion kalman filter for discrete multichannel arma signals, in: Proceedings of the 2003 IEEE International symposium on intelligent control, IEEE, 2003, pp. 377--382.
\newblock \href {https://doi.org/10.1109/ISIC.2003.1254663} {\path{doi:10.1109/ISIC.2003.1254663}}.

\bibitem{yoon2023adaptive}
K.~Yoon, J.~Choi, K.~Huh, Adaptive decentralized sensor fusion for autonomous vehicle: Estimating the position of surrounding vehicles, IEEE Access 11 (2023) 90999--91008.
\newblock \href {https://doi.org/10.1109/ACCESS.2023.3308152} {\path{doi:10.1109/ACCESS.2023.3308152}}.

\bibitem{mukherjee2022resilient}
M.~Mukherjee, A.~Banerjee, G.~Nikolakopoulos, A resilient to faults auto-encoder enabled kalman based multi-sensorial fusion, in: 2022 30th Mediterranean Conference on Control and Automation (MED), IEEE, 2022, pp. 662--669.
\newblock \href {https://doi.org/10.1109/MED54222.2022.9837222} {\path{doi:10.1109/MED54222.2022.9837222}}.

\bibitem{mukherjee2023fault}
M.~Mukherjee, A.~Banerjee, S.~G. Satpute, G.~Nikolakopoulos, Fault resilient decentralized multi-sensorial fusion based pose estimation for autonomous navigation around asteroid, International Journal of Control, Automation and Systems 21~(6) (2023) 2031--2042.
\newblock \href {https://doi.org/https://doi.org/10.1007/s12555-022-0528-3} {\path{doi:https://doi.org/10.1007/s12555-022-0528-3}}.

\bibitem{mukherjee2022auto}
M.~Mukherjee, A.~Banerjee, G.~Nikolakopoulos, An auto-encoder enabled fault detection and isolation scheme for enabling a multi-sensorial distributed pose estimation, in: 2022 IEEE 31st International Symposium on Industrial Electronics (ISIE), IEEE, 2022, pp. 281--288.
\newblock \href {https://doi.org/10.1109/ISIE51582.2022.9831458} {\path{doi:10.1109/ISIE51582.2022.9831458}}.

\bibitem{mukherjee2024autonomous}
M.~Mukherjee, A.~Banerjee, A.~Koval, G.~Nikolakopoulos, Autonomous residual threshold detection for decentralized radar inertial odometry in hostile environments, in: 2024 International Conference on Control, Automation and Diagnosis (ICCAD), IEEE, 2024, pp. 1--7.
\newblock \href {https://doi.org/10.1109/ICCAD60883.2024.10553935} {\path{doi:10.1109/ICCAD60883.2024.10553935}}.

\bibitem{mukherjee2023decentralized}
M.~Mukherjee, A.~Banerjee, A.~Koval, G.~Nikolakopoulos, Decentralized fusion-based ego velocity estimation using multiple fmcw radars, in: 2023 21st International Conference on Advanced Robotics (ICAR), IEEE, 2023, pp. 244--251.
\newblock \href {https://doi.org/10.1109/ICAR58858.2023.10406467} {\path{doi:10.1109/ICAR58858.2023.10406467}}.

\bibitem{sun2004multi}
S.-L. Sun, Z.-L. Deng, Multi-sensor optimal information fusion kalman filter, Automatica 40~(6) (2004) 1017--1023.
\newblock \href {https://doi.org/https://doi.org/10.1016/j.automatica.2004.01.014} {\path{doi:https://doi.org/10.1016/j.automatica.2004.01.014}}.

\bibitem{noack2017decentralized}
B.~Noack, J.~Sijs, M.~Reinhardt, U.~D. Hanebeck, Decentralized data fusion with inverse covariance intersection, Automatica 79 (2017) 35--41.
\newblock \href {https://doi.org/https://doi.org/10.1016/j.automatica.2017.01.019} {\path{doi:https://doi.org/10.1016/j.automatica.2017.01.019}}.

\bibitem{pirmoradi2009fault}
F.~Pirmoradi, F.~Sassani, C.~De~Silva, Fault detection and diagnosis in a spacecraft attitude determination system, Acta Astronautica 65~(5-6) (2009) 710--729.
\newblock \href {https://doi.org/https://doi.org/10.1016/j.actaastro.2009.03.002} {\path{doi:https://doi.org/10.1016/j.actaastro.2009.03.002}}.

\bibitem{yuan2021fault}
Z.~Yuan, N.~Song, X.~Pan, J.~Song, F.~Ma, Fault detection, isolation, and reconstruction for satellite attitude sensors using an adaptive hybrid method, IEEE Transactions on Instrumentation and Measurement 70 (2021) 1--12.
\newblock \href {https://doi.org/10.1109/TIM.2021.3097404} {\path{doi:10.1109/TIM.2021.3097404}}.

\bibitem{chafouk2011fault}
H.~Chafouk, et~al., Fault detection and isolation using kalman filter bank for a wind turbine generator, in: 2011 19th Mediterranean Conference on Control \& Automation (MED), IEEE, 2011, pp. 144--149.
\newblock \href {https://doi.org/10.1109/MED.2011.5983155} {\path{doi:10.1109/MED.2011.5983155}}.

\bibitem{koval2022dataset}
A.~Koval, S.~Karlsson, S.~S. Mansouri, C.~Kanellakis, I.~Tevetzidis, J.~Haluska, A.-a. Agha-mohammadi, G.~Nikolakopoulos, Dataset collection from a subt environment, Robotics and Autonomous Systems 155 (2022) 104168.
\newblock \href {https://doi.org/https://doi.org/10.1016/j.robot.2022.104168} {\path{doi:https://doi.org/10.1016/j.robot.2022.104168}}.

\bibitem{kyuroson2023multimodal}
A.~Kyuroson, N.~Dahlquist, N.~Stathoulopoulos, V.~K. Viswanathan, A.~Koval, G.~Nikolakopoulos, Multimodal dataset from harsh sub-terranean environment with aerosol particles for frontier exploration, in: Mediterranean Conference on Control and Automation (MED2023), 2023.
\newblock \href {https://doi.org/10.1109/MED59994.2023.10185906} {\path{doi:10.1109/MED59994.2023.10185906}}.

\end{thebibliography}

\end{document}